\documentclass{article} % For LaTeX2e
\usepackage{iclr2027_conference,times}

\usepackage{amsmath,amsfonts,bm}

\def\eqref#1{equation~\ref{#1}}
\def\1{\bm{1}}

\DeclareMathAlphabet{\mathsfit}{\encodingdefault}{\sfdefault}{m}{sl}
\SetMathAlphabet{\mathsfit}{bold}{\encodingdefault}{\sfdefault}{bx}{n}

\newcommand{\KL}{D_{\mathrm{KL}}}

\usepackage{multirow}
\usepackage{url}
\usepackage{graphicx}
\usepackage{enumitem}
\usepackage{etoolbox}
\usepackage{booktabs}
\usepackage{wrapfig}
\usepackage{float}
\usepackage{hyperref}
\AtBeginEnvironment{table}{\setlength{\belowcaptionskip}{10pt}}
\AtBeginEnvironment{table*}{\setlength{\belowcaptionskip}{10pt}}
\newcommand{\student}{\pi_{\theta}}
\newcommand{\teacher}{\pi_{\phi}}

\title{When Should the Teacher Move? \\
Temporal Coupling and Stability in Self On-Policy Distillation
\vspace {0.7em} }
\author{
\begin{tabular}{@{}c@{}}
\makebox[\dimexpr\textwidth-2\tabcolsep-3pt\relax][c]
{\scalebox{0.95}{\bfseries 
Haowei Guo$^{1,*}$,\hspace{0.35em}% 
Baolong Bi$^{2,*}$,\hspace{0.35em}% 
Ruicheng Zhang$^{3}$,\hspace{0.35em}% 
Cong Li$^{1,\dagger}$,\hspace{0.35em}% 
Bingqian Sun$^{1}$,\hspace{0.35em}% 
Zhonghai Wu$^{1,\dagger}$}} \\[0.2em]
{\normalfont\small 
$^1$Peking University\hspace{1.2em}% 
$^2$University of Chinese Academy of Sciences\hspace{1.2em}% 
$^3$Tsinghua University } \\[0.12em]
{\normalfont\small
\texttt{guohaowei26@stu.pku.edu.cn},\hspace{0.7em}%
\texttt{\{li.cong,wuzh\}@pku.edu.cn}} \\[0.08em]
{\normalfont\footnotesize 
$^*$Equal contribution.\hspace{1em}% 
$^\dagger$Corresponding authors.}
\end{tabular}
}

\iclrfinalcopy

\begin{document}

\maketitle
\thispagestyle{plain}
\pagestyle{plain}
%%%%%%%%%%%%%%%%%%%%%%%%%%%%%%%%%%%%%%%%%%%%%%%%%%%%%%%%%%%
\begin{abstract}
%%%%%%%%%%%%%%%%%%%%%%%%%%%%%%%%%%%%%%%%%%%%%%%%%%%%%%%%%%%

Self on-policy distillation trains a student policy against a teacher
derived from its own parameter history, yet the teacher's update
schedule---which governs the \emph{temporal coupling} between teacher
and student---has not been systematically studied as a stability
variable. Through a controlled schedule sweep on Qwen3-8B, we
establish that \emph{isolation periods}---complete teacher freezing
between updates---are the key structural property enabling stable
learning, not teacher age, and we introduce a diagnostic framework
of temporal KL structure, refresh shock, and length-tail risk to
characterize training dynamics. We further identify
\emph{state-oblivious collapse}: the best short-horizon fixed
schedule ($M{=}50$) catastrophically fails under long-horizon training
because a clock-driven refresh can copy a transiently drifting student
into the teacher in a single, irreversible step---a failure mode
invisible under short-horizon evaluation and mechanistically distinct
from EMA's chronic contamination. To address this, we propose
\emph{Consolidation-Gated Teacher Refresh} (CGTR), which preserves
isolation periods while gating each refresh on joint evidence of
reward improvement and length-tail safety, ensuring every teacher
movement responds to genuine student consolidation rather than
a clock signal. With a single shared parameter set and no per-dataset
retuning, CGTR achieves \textbf{zero collapse} and the best final
score on all four tasks (Chemistry, Biology, Physics, ToolUse),
self-regulating its refresh frequency to each task's learning dynamics. 
\end{abstract}

%%%%%%%%%%%%%%%%%%%%%%%%%%%%%%%%%%%%%%%%%%%%%%%%%%%%%%%%%%%
\section{Introduction}
\label{sec:intro}
%%%%%%%%%%%%%%%%%%%%%%%%%%%%%%%%%%%%%%%%%%%%%%%%%%%%%%%%%%%

\begin{wrapfigure}{r}{0.48\textwidth}
  \vspace{-0.8\baselineskip}
  \centering
  \includegraphics[width=\linewidth]{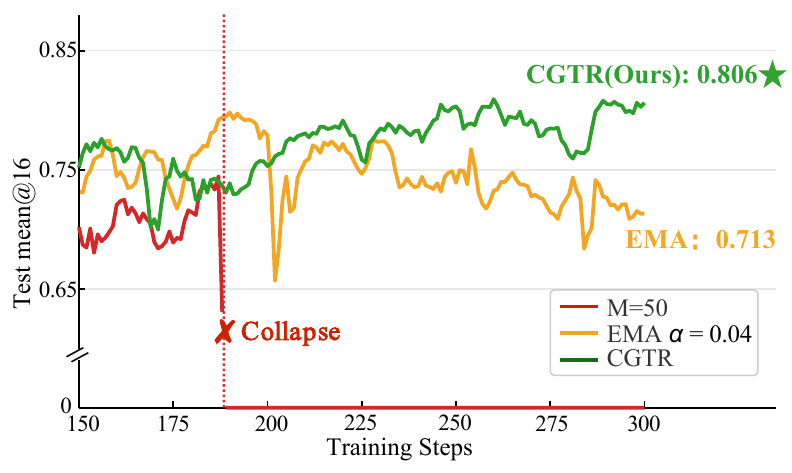}
  \vspace{-1\baselineskip}
  \caption{%
    Long-horizon results (300 steps, Chemistry, steps 150--300 shown).
    $M{=}50$ collapses catastrophically at step~189 (test mean@16
    $\to 0$); EMA $\alpha{=}0.04$ plateaus at $0.713$;
    \textbf{CGTR (ours) achieves the best final score of $\mathbf{0.806}$}
    with zero collapse.
  }
  \label{fig:overview}
  \vspace{-0.6\baselineskip}
\end{wrapfigure}

Large language model post-training increasingly relies on on-policy
learning~\citep{schulman2017proximal,ouyang2022training,shao2024deepseekmath,guo2025deepseekr1},
where the model improves by sampling from its own
policy rather than imitating static demonstrations.
On-policy distillation extends this principle by providing the student
with dense token-level supervision from a teacher during on-policy
rollouts \citep{agarwal2024onpolicy}, reducing the distribution mismatch
inherent in classical off-policy distillation.
A natural extension is to derive this teacher from the model itself
rather than an external source, as recent self-distillation methods do
by constructing teacher signals from the same model under richer
contexts---privileged reasoning, environment feedback, or
compression-oriented instructions
\citep{zhao2026selfdistilled,hubotter2026reinforcement,
li2026rethinking,yang2026selfdistilledrlvr,sang2026crisp}.

Unlike a fixed external teacher, a self-teacher co-evolves with the
student it trains, introducing a fundamental tension that prior work has
not systematically studied.
On the one hand, the teacher must follow the student over time to
support bootstrapping and remain relevant.
On the other hand, if the teacher parameters track the student too
closely, it loses \emph{temporal independence}, weakening the corrective
distillation signal because the teacher no longer retains a historical
parameter state capable of counteracting student drift.
The teacher must therefore serve as a robust anchor capable of
rescuing the student from suboptimal distributions---not a passive copy
that ratifies whatever the student does.
We call the degree to which the teacher tracks the student
\emph{temporal coupling}, and ask: \emph{how does temporal coupling
shape training stability and long-horizon learning?}

Each canonical mechanism for controlling temporal coupling embodies a
distinct failure mode, but prior work treats these mechanisms as
implementation choices rather than stability-critical variables.
Tight coupling (e.g., updating the teacher at every step) removes the
temporal separation between teacher and student parameters, causing
\emph{reference collapse}: the corrective signal from a historical
teacher vanishes, leaving training unconstrained against
self-reinforcing student drift.
EMA \citep{polyak1992acceleration,tarvainen2017mean,grill2020bootstrap}
mitigates this by diluting each update, but continuous soft tracking
means that student drift accumulates in the teacher without bound over
long training horizons, leading to \emph{chronic teacher contamination}.
Sparse hard refresh \citep{mnih2015human} avoids both failure modes in
the short run: by periodically copying the student into the teacher and
then freezing it for many steps, creating \emph{isolation periods} during
which student fluctuations cannot contaminate the teacher.
These isolation periods help preserve temporal independence by preventing
the teacher from immediately tracking student fluctuations.  Our 100-step
schedule sweep attributes their stability benefit to this effect rather
than teacher age alone (Section~\ref{sec:results-m50}), with long-horizon
experiments providing further evidence (Figure~\ref{fig:overview}).

Yet isolation periods alone are not sufficient for long-horizon stability
when the refresh schedule is fixed.
A fixed interval is oblivious to student state: the teacher is
updated at every $M$-th step regardless of whether the student is consolidating or drifting.
If a hard refresh fires at the wrong moment (e.g., during a transient
entropy spike or a reward plateau), the teacher absorbs corrupted
behavior in a single, irreversible copy.
Once copied, the contaminated teacher continuously reinforces the
corrupted state through subsequent distillation, and there is no
partial recovery path thereafter.

This analysis motivates a schedule that preserves the structural
isolation advantage of hard refresh while conditioning each refresh on
observable evidence that the student has genuinely consolidated.
We propose \emph{Consolidation-Gated Teacher Refresh} (CGTR), which
maintains a minimum isolation period and gates each refresh on the joint satisfaction of a reward-improvement condition
and a length-tail safety condition.
When all conditions are met, the refresh is corrective; when any
fails, the teacher remains frozen regardless of elapsed time.
This design eliminates the need to pre-specify a refresh interval while
ensuring that every teacher movement is strictly a response to genuine
progress rather than a response to a clock signal alone.

Our contributions are as follows:
\vspace{-0.2em}
\begin{itemize}[leftmargin=1.6em,itemsep=2pt,topsep=3pt]
  \item We identify \emph{isolation period} (complete teacher freezing
    between updates) as the key structural property enabling stable
    self on-policy distillation, supported by a systematic schedule sweep
    and a diagnostic framework of temporal KL, refresh shock, and
    length-tail risk.
  \item We characterize \emph{state-oblivious collapse}, a long-horizon failure mode of fixed-interval hard refresh in which a single ill-timed update causes irreversible capability
    loss invisible at short training horizons.
  \item We propose \emph{Consolidation-Gated Teacher Refresh} (CGTR),
    which gates each refresh on reward improvement and length-tail safety.
    CGTR achieves zero collapse and the best score on all four tasks
    (Chemistry, Biology, Physics, ToolUse) with a single parameter set.
\vspace{-0.2em}
\end{itemize}

%%%%%%%%%%%%%%%%%%%%%%%%%%%%%%%%%%%%%%%%%%%%%%%%%%%%%%%%%%
\section{Related Work}
\label{sec:related}
%%%%%%%%%%%%%%%%%%%%%%%%%%%%%%%%%%%%%%%%%%%%%%%%%%%%%%%%%%%

\paragraph{On-Policy Distillation and Self-Distillation.}
Knowledge distillation trains a student to match a teacher's predictive
distribution~\citep{hinton2015distilling}, which sequence-level
distillation adapts for autoregressive generation by having the student
imitate teacher-generated sequences~\citep{kim-rush-2016-sequence}.  In
reinforcement learning, policy distillation transfers behavior between
teacher and student policies~\citep{rusu2015policy}, and on-policy
distillation extends this idea by training the student on its own
sampled trajectories with dense teacher supervision~\citep{agarwal2024onpolicy}.
Iterative self-training showed that a model can bootstrap its
own training signal by generating and filtering self-produced
reasoning~\citep{zelikman2022star}.
Recent self-distillation methods further remove the need for a stronger
external teacher by deriving teacher signals from the same model under
richer contexts, such as privileged reasoning information, environment
feedback, reward-verifiable directions, or compression-oriented
instructions
\citep{zhao2026selfdistilled,hubotter2026reinforcement,
yang2026selfdistilledrlvr,sang2026crisp}.  These methods demonstrate
that self-teachers can provide dense supervision for post-training
without a separate teacher model.  Our work is orthogonal: rather than
studying how to construct the self-teacher signal, we study how the
self-teacher should move over training time.

\paragraph{Stability and Failure Modes in Self-Distillation.}
Recent work has begun to analyze why self-distillation succeeds or fails.
\citet{li2026rethinking} show that teacher--student compatibility and
thinking-pattern alignment critically affect distillation outcomes.
Other works identify information leakage, unstable long-term training,
and the need for reward-grounded or periodically synchronized teachers
as key concerns~\citep{yang2026selfdistilledrlvr,sang2026crisp}.
These studies treat teacher movement as a fixed implementation choice,
and different mechanisms---hard refresh, fixed teachers, and
EMA~\citep{mnih2015human,polyak1992acceleration,tarvainen2017mean}---are
adopted without analyzing whether the optimal interval is task-dependent.
We address this gap by treating temporal coupling as a first-class
stability variable, identifying what structural property enables
long-horizon consolidation, and showing that the optimal hard refresh
interval varies with task characteristics---motivating CGTR as an
adaptive schedule that generalizes across settings.

%%%%%%%%%%%%%%%%%%%%%%%%%%%%%%%%%%%%%%%%%%%%%%%%%%%%%%%%%%%
\section{Analysis of Teacher Update Schedules}
\label{sec:analysis}
%%%%%%%%%%%%%%%%%%%%%%%%%%%%%%%%%%%%%%%%%%%%%%%%%%%%%%%%%%%

% Stable self on-policy distillation requires the teacher to remain a
% corrective reference rather than a passive copy of the student.
% We investigate how the teacher update schedule shapes training stability
% and long-horizon learning in self on-policy distillation.
% Section~\ref{sec:setup} establishes the formal setup and the two canonical
% update rules.
% Sections~\ref{sec:results-collapse}--\ref{sec:analysis-fragility}
% then reveal qualitatively distinct failure modes through a systematic
% schedule sweep on Qwen3-8B / SciKnowEval Chemistry, and
% Section~\ref{sec:analysis-summary} consolidates the structural requirements
% that emerge from this sweep.

\subsection{Preliminary: Setup and Update Rules}
\label{sec:setup}

% Let $\student$ denote the \emph{student policy} parameterized by
% $\theta$, trained via gradient updates on a reward signal and a
% distillation objective.  Let $\teacher$ denote the \emph{self-teacher
% policy} parameterized by $\phi$, which is derived from a historical
% version of $\theta$.  The combined training objective at step $t$ is:
% \vspace{-0.1mm}
% \begin{equation}
% \mathcal{L}_t(\theta) = -\mathbb{E}_{x \sim \student}\bigl[r(x)\bigr]
% + \lambda \cdot \KL\bigl(\student \,\|\, \teacher\bigr),
% \label{eq:objective}
% \end{equation}
% \vspace{-0.5mm}
% where $r(\cdot)$ is the task reward and $\lambda$ controls the strength
% of the distillation regularizer. In practice, the reward term uses GRPO-style policy gradient and the
% KL term is approximated by a top-$k$ Jensen--Shannon Divergence (JSD) with importance-sampling
% correction (Appendix~\ref{sec:appendix-repro}).

Let $\student$ denote the \emph{student policy} parameterized by
$\theta$, and let $\teacher$ denote the \emph{self-teacher policy}
parameterized by $\phi$, which is derived from a historical version
of $\theta$.  For a student rollout $y$ and feedback context $f$, we
optimize:
\vspace{-2.5mm}
\begin{equation}
\mathcal{L}_t(\theta) =
\sum_{j=1}^{|y|}\mathrm{JSD}\!\left(
\pi_\theta(\cdot\mid x,y_{<j})\,\|\,\operatorname{stopgrad}
\pi_\phi(\cdot\mid x,f,y_{<j})\right),
\label{eq:objective}
\vspace{-3.5mm}
\end{equation}

where $f$ is the feedback context derived from rollout outcomes.
We approximate JSD with top-$k$ logits and importance-sampling
correction (Appendix~\ref{sec:appendix-repro}).

Two canonical update rules define the temporal coupling between $\phi$
and $\theta$:

\vspace{-0.5em}
\paragraph{Periodic hard refresh.}
The teacher parameters are updated to match the student every $M$ steps:
\vspace{-0.75mm}
\begin{equation}
\phi_t =
\begin{cases}
\theta_t & \text{if } t \bmod M = 0, \\
\phi_{t-1} & \text{otherwise.}
\end{cases}
\label{eq:hard-refresh}
\vspace{-2.85mm}
\end{equation}

Between consecutive refreshes, $\teacher$ is completely frozen, forming
an \emph{isolation period} during which student fluctuations cannot
contaminate the teacher.

\vspace{-0.5em}
\paragraph{Exponential moving average (EMA).}
The teacher is updated continuously at every step:
\vspace{-0.25mm}
\begin{equation}
\phi_t = (1 - \alpha)\,\phi_{t-1} + \alpha\,\theta_t,
\quad 0 < \alpha \le 1.
\label{eq:ema}
\end{equation}
EMA has no isolation period: the teacher absorbs a fraction $\alpha$
of student updates at every step, bounding per-step contamination
but allowing drift to accumulate over long horizons.

To control for teacher age, we derive a lag-aligned EMA baseline
($\alpha{=}0.04$; Appendix~\ref{sec:appendix-lag}).
Table~\ref{tab:main} summarizes the schedule sweep results (setup in
Section~\ref{sec:exp-setup}; terminology in
Appendix~\ref{sec:appendix-concepts}).

\begin{table*}[t]
\caption{%
Schedule sweep results on Qwen3-8B / SciKnowEval Chemistry (100 steps).
\textbf{Test\,m@16}: test mean@16 accuracy over 16 samples (primary);
\textbf{B@4}: bootstrap-estimated best-of-4 test accuracy;
\textbf{Ent.}: token entropy (training mean / std);
\textbf{Rew.}: mean rollout reward;
\textbf{KL}: mean student--teacher KL;
\textbf{Collapse}: denotes the mechanism
discussed in Section~\ref{sec:results-collapse}(--- = none observed).
$^\ddagger$M1 terminated at step~16; no test score available.
Bold marks the highest test scores only.
}
\centering
\small
\setlength{\tabcolsep}{4pt}
\begin{tabular}{lccccccc}
\toprule
\textbf{Schedule} &
  \textbf{Test\,m@16} &
  \textbf{B@4} &
  \textbf{Ent.\,mean} &
  \textbf{Ent.\,std} &
  \textbf{Rew.\,mean} &
  \textbf{KL\,mean} &
  \textbf{Collapse} \\
\midrule
$M=1$$^\ddagger$ & N/A    & N/A    & $\sim$1.05 & ---   & ---   & ---   & step 10 \\
$M=2$    & 0.225  & 0.276  & 0.900 & 1.063 & 0.504 & 0.031 & step 87 \\
$M=3$    & 0.673  & 0.740  & 1.715 & 1.167 & 0.607 & 0.085 & --- \\
$M=4$    & 0.656  & 0.718  & 1.536 & 1.131 & 0.595 & 0.053 & --- \\
$M=5$    & 0.653  & 0.723  & 2.001 & 1.091 & 0.629 & 0.064 & --- \\
$M=8$    & 0.596  & 0.629  & 1.413 & 0.713 & 0.597 & 0.004 & --- \\
$M=10$   & 0.669  & 0.737  & 0.545 & 0.278 & 0.614 & 0.025 & --- \\
$M=20$   & 0.629  & 0.689  & 0.595 & 0.119 & 0.598 & 0.004 & --- \\
$M=50$   & \textbf{0.755} & \textbf{0.827} & 0.499 & 0.048 & 0.629 & 0.005 & --- \\
\midrule
Fixed    & 0.697  & 0.751  & 0.437 & 0.097 & 0.637 & 0.008 & --- \\
\midrule
EMA\,$\alpha$ = 0.1  & 0.645 & 0.689 & 1.417 & 0.694 & 0.613 & 0.005 & --- \\
EMA\,$\alpha$ = 0.04 & 0.713 & 0.782 & 0.727 & 0.142 & 0.665 & 0.022 & --- \\
\bottomrule
\end{tabular}
\label{tab:main}

\end{table*}

\begin{figure*}[t]
\centering
\includegraphics[width=\linewidth]{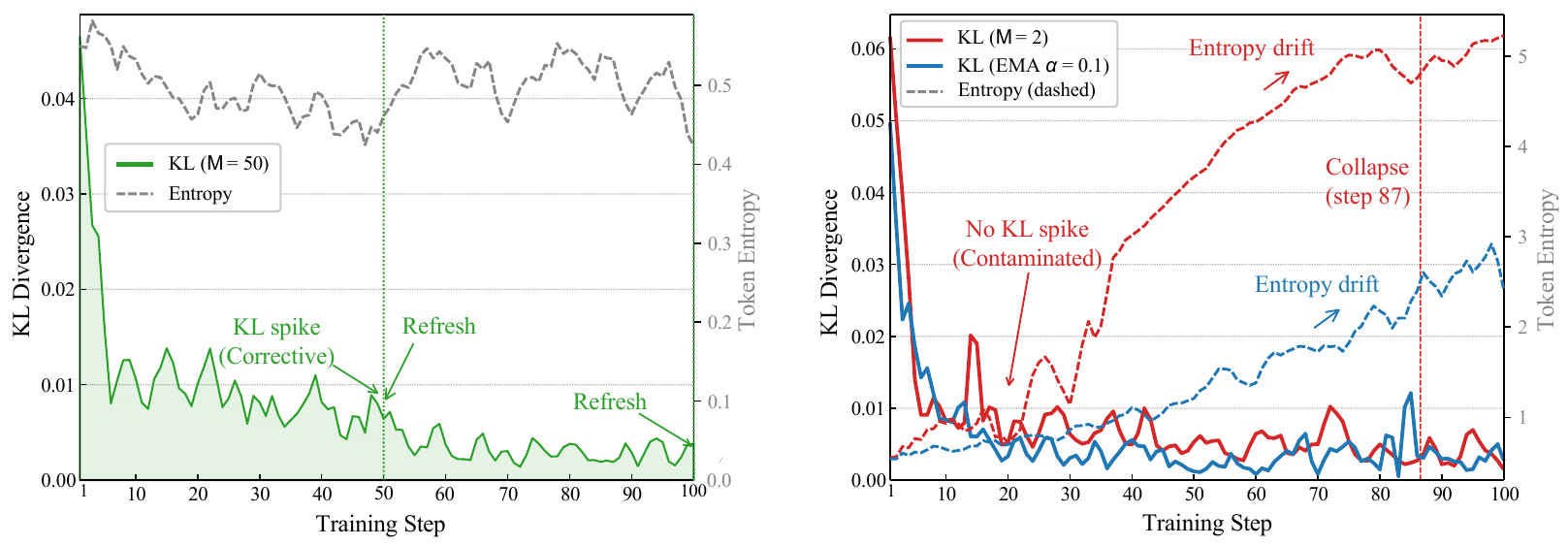}
\vspace{-4mm}
\caption{%
Illustration of two refresh dynamics. \textbf{(Left)}~A \emph{corrective refresh}
(e.g., $M{=}50$) produces a transient KL spike as the teacher absorbs
genuine improvement; the student then re-adapts under the new target.
\textbf{(Right)}~A \emph{contaminating} pattern (low $M$ or EMA with high $\alpha$)
produces no correction spike and is followed by gradual entropy growth
as student drift propagates through the teacher.
}
\label{fig:refresh-shock}
\vspace{-0.8em}
\end{figure*}

\subsection{Fast Synchronization Causes Reference Collapse}
\label{sec:results-collapse}

% When the teacher is updated at every step ($M{=}1$), the teacher and
% student parameters are virtually identical. The KL divergence
% $\KL(\student \| \teacher)$ approaches zero regardless of the
% student's behavior, which means the regularizer effectively
% disappears. Without any corrective signal, sequence lengths begin to
% grow unchecked: normalized sequence length exceeds 8,000 at step~10 and
% reaches 16,908 at step~14, forcing early termination at step~16.
% No test score is available.

% For $M{=}2$, the teacher lags by at most one step, which still provides
% only negligible temporal independence. Training proceeds normally
% until step~87, at which point a rapid collapse unfolds: entropy drops
% precipitously to 0.047 (versus a pre-collapse mean near 0.5)
% while normalized sequence length rises to 15,126.  The final test
% score after collapse is 0.225. The pattern is consistent with a
% \emph{delayed reference collapse}: the teacher has been silently
% tracking the student's lengthening trajectories, so when the student
% reaches a critical instability point, it provides no
% corrective pressure.

% The key mechanism is \emph{loss of temporal independence}, not simply
% high KL divergence.  Indeed, the KL mean for $M{=}2$ is only 0.031
% --- a seemingly small value.  But this low KL reflects not strong
% constraint but rather the collapse of constraint: the teacher has
% already matched the student's drifted distribution, so the KL
% penalty is satisfied trivially.

At $M{=}1$, teacher and student parameters are virtually identical,
eliminating temporal independence and the corrective signal of a
historical teacher.  Without this anchor, sequence lengths grow
unchecked, exceeding 8,000 at step~10 and reaching 16,908 by step~14,
forcing early termination with no test score.
At $M{=}2$, the teacher lags by only one step, providing negligible
temporal independence.  Around the reference-collapse marker
at step~87, mean response length rises sharply toward the 8,192-token
cap (Figure~\ref{fig:length-tail});
the final test score is 0.225.
The diagnostic KL mean at $M{=}2$ is only 0.031, consistent with tight coupling
and a weak historical parameter constraint rather than a strong
independent anchor.
This \emph{loss of temporal independence} underlies both
failures---a pattern we term \emph{reference collapse}.
Accuracy-based collapse is assessed separately
(Appendix~\ref{sec:collapse-detection}).

\paragraph{Diagnostic: KL temporal structure.}
What matters is not the KL level but its \emph{temporal structure}: a
corrective refresh produces a transient KL spike as the student
re-adapts to the new teacher; a contaminating pattern shows no spike
but subsequent entropy growth.
Figure~\ref{fig:refresh-shock} contrasts these two patterns,
providing a diagnostic that distinguishes healthy from contaminating
training dynamics.

\subsection{Sparse Hard Refresh Achieves Best Short-Horizon Performance}
\label{sec:results-m50}

Among all schedule-sweep settings, $M{=}50$ achieves the best test score
($0.755$ at step~100), with entropy mean $0.499$ and KL mean $0.005$.
These results confirm that
isolation periods---the complete freezing of the teacher between
updates---are the key mechanism for stability and performance.

The advantage of $M{=}50$ over EMA $\alpha{=}0.04$ is not attributable
to teacher age: lag-aligned EMA ($\alpha{=}0.04$;
Appendix~\ref{sec:appendix-lag}) matches $M{=}50$ in average teacher
age yet produces markedly worse entropy stability (mean 0.727 vs.\
0.499).  The difference is structural: each isolation window
completely freezes the teacher, whereas EMA incorporates a fraction
of student updates at every step and \emph{never resets the
teacher-student gap}---drift accumulates unboundedly regardless of
the per-step mixing rate---a failure mode we term
\emph{steady-state contamination}.
This establishes a key principle: \emph{isolation period, not teacher
age, enables stable self-distillation}.
Furthermore, entropy alone is an unreliable consolidation signal:
it can be high during productive exploration or suppressed by
over-regularization.
Reward improvement directly verifies that the student
has made genuine progress.

\subsection{The Fragility of Fixed Isolation Periods}
\label{sec:analysis-fragility}

The schedule-sweep results reveal a second structural observation:
while sparse hard refresh creates the best short-horizon stability,
a fixed update interval is inherently \emph{oblivious to student state}.
The teacher is refreshed at every $M$-th step regardless of whether the
student is currently consolidating or drifting.  If a refresh fires
during a transient entropy spike or a reward plateau, the teacher absorbs
% corrupted behavior in a single, irreversible copy.
% Unlike EMA, which limits contamination per step by $\alpha$, a hard copy
% applies the full student state instantaneously---with no partial recovery
% path once the teacher has converged to the drifted student distribution.
corrupted behavior in a single, irreversible copy---unlike EMA, which
limits contamination per step by $\alpha$, a hard copy applies the full
student state instantaneously with no partial recovery path.
The longer the isolation period, the larger the potential drift that can
accumulate before the next refresh, amplifying this irreversibility risk
in proportion to the very property that made isolation beneficial---a
failure mode we term \emph{state-oblivious collapse}.
% This implies a fundamental tension: longer isolation periods provide
% better short-horizon stability, but they also increase the risk that
% when the isolation ends the student is in a temporarily unstable state.
% Because the schedule fires at every $M$-th step regardless of student
% state, any transient drift present at that moment is captured fully
% and irreversibly---there is no partial recovery path once the teacher
% has converged to the drifted distribution.  The longer the isolation
% period, the larger the potential drift that can accumulate before the
% next refresh, amplifying this irreversibility risk in proportion to
% the very property that made isolation beneficial.

Compounding this, the optimal interval $M^*$ is dataset-dependent:
different tasks have different learning dynamics, reward densities, and
trajectory length characteristics, so the collapse boundary shifts
across datasets.  A single pre-specified $M$ is therefore unlikely to
be simultaneously safe across all tasks.
This motivates an adaptive refresh rule that preserves isolation periods
while conditioning each update on whether the student has genuinely
consolidated, which we develop in Section~\ref{sec:cgtr}.

\paragraph{Diagnostic: length-tail expansion as an early warning.}
Length-tail expansion can appear before abrupt changes in entropy or
sequence length become visible in mean values. Figure~\ref{fig:length-tail} illustrates how tail expansion in $M{=}2$
precedes the abrupt changes at step~87 by roughly 12 steps.
Monitoring response length tail ratio growth provides a fast, stateless
early-warning signal, separate from the accuracy-based collapse criterion.
%; we incorporate this as the length-tail gate in CGTR.

\begin{table}[t]
\caption{Two design requirements for stable teacher update,
validated in Sections~\ref{sec:results-collapse}--\ref{sec:analysis-fragility}.
Teacher isolation indicates whether the teacher remains fully frozen for
a sufficient interval; state-aware refresh indicates whether an update
depends on the observed student state.}
\centering
\small
\renewcommand{\arraystretch}{1.08}
\setlength{\tabcolsep}{3.2pt}
\begin{tabular*}{0.8\linewidth}{@{\extracolsep{\fill}}lccc@{}}
\toprule
\raisebox{0.8ex}{\textbf{Schedule}} &
  \shortstack{\textbf{Teacher}\\\textbf{isolation}} &
  \shortstack{\textbf{State-aware}\\\textbf{refresh}} &
  \shortstack{\textbf{Observed}\\\textbf{failure}} \\
\midrule
Tight refresh ($M{\le}2$) &
  Insufficient & No & Reference collapse \\
EMA  &
  None & No & Steady-state contamination \\
Fixed refresh ($M$ large) &
  Sufficient & No & State-oblivious collapse \\
\midrule
\textbf{CGTR (ours)} & Sufficient & Yes & --- \\
\bottomrule
\end{tabular*}
% \vspace{-2mm}
\label{tab:requirements}
\vspace{-3.6mm}
\end{table}

\subsection{Two Design Requirements}
\label{sec:analysis-summary}

The schedule sweep above reveals two design requirements that any
viable teacher update rule must satisfy simultaneously.

\textbf{\emph{(A) Sufficient teacher isolation.}}
The teacher must remain frozen long enough to avoid immediately absorbing
student fluctuations. This isolation helps preserve temporal independence:
it prevents student drift from immediately entering the teacher,
preserving a corrective historical reference. Tight hard refresh
($M{\le}2$) provides too little isolation, while EMA provides none.

\textbf{\emph{(B) State-aware refresh.}}
Each teacher update must be triggered by evidence that the student
has consolidated, rather than by elapsed training steps alone.
A refresh that fires during transient drift copies the corrupted
state into the teacher in a single, irreversible step.
Clock-based schedules violate this requirement by construction.

Table~\ref{tab:requirements} maps each canonical schedule to these
% two requirements and the observed failure mode when a requirement is
% violated.
% No clock-based schedule satisfies both simultaneously:
% tight coupling destroys parameter independence; EMA lacks any isolation
% period and absorbs student drift at every step; and fixed-interval hard
% refresh, while providing isolation, fires without regard to student state.
% An effective update rule must therefore combine structural isolation with
% observable evidence of student consolidation.
requirements; no clock-based schedule satisfies both,
motivating a rule that combines sufficient teacher isolation with
state-aware refreshes.

%%%%%%%%%%%%%%%%%%%%%%%%%%%%%%%%%%%%%%%%%%%%%%%%%%%%%%%%%%%
\section{Consolidation-Gated Teacher Refresh}
\label{sec:cgtr}
%%%%%%%%%%%%%%%%%%%%%%%%%%%%%%%%%%%%%%%%%%%%%%%%%%%%%%%%%%%
\begin{figure}[t]
  \centering
  \includegraphics[width=0.98\columnwidth]{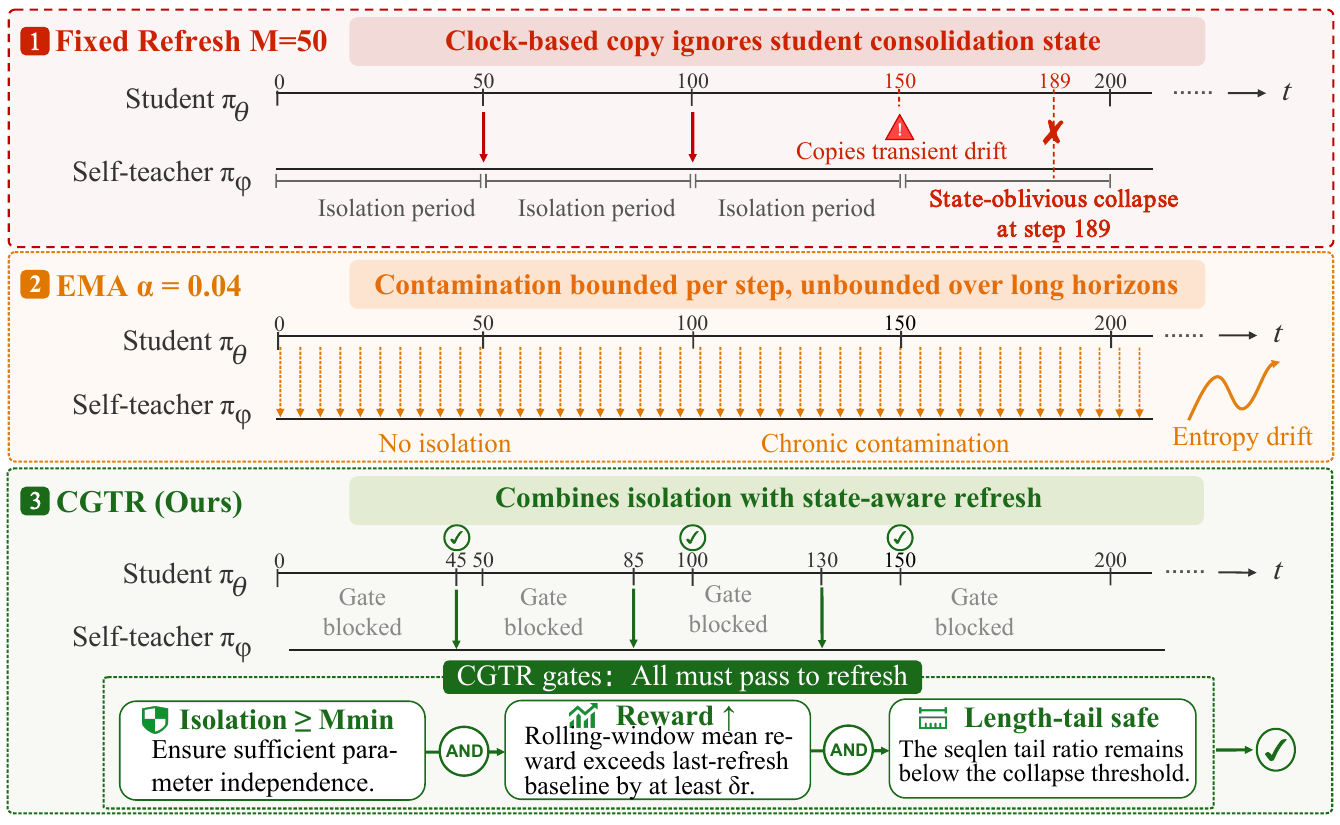}
  \vspace{-2mm}
  \caption{Comparison of teacher-update strategies.
    Fixed refresh ignores student state, EMA continuously tracks student
    updates, and CGTR preserves isolation while refreshing only when the
    isolation, reward, and length-tail gates jointly pass.
    % Comparison of three teacher update strategies.
    % \textbf{(1) Fixed refresh ($M{=}50$):} clock-based copies create
    % isolation periods but ignore student state, causing state-oblivious
    % collapse when a refresh fires during transient drift (step~189).
    % \textbf{(2) EMA ($\alpha{=}0.04$):} no isolation period; contamination
    % is bounded per step but accumulates unboundedly over long horizons.
    % \textbf{(3) CGTR (ours):} combines isolation periods with
    % consolidation-gated refreshes---all three conditions (isolation,
    % reward, length-tail) must pass before the teacher moves.
  }
  \label{fig:cgtr-method}
  \vspace{-0.37em}
\end{figure}

\emph{Consolidation-Gated Teacher Refresh} (CGTR) is designed to
satisfy both design requirements from
Section~\ref{sec:analysis-summary} simultaneously: a hard isolation
floor provides sufficient teacher isolation, while conditional gating
on observable consolidation signals makes each refresh state-aware.
Each condition can be evaluated from a single rolling window with no
additional model inference or learned component
(Figure~\ref{fig:cgtr-method}).
The full update rule is:
\begin{equation}
\phi_t =
\begin{cases}
\theta_t   & \text{if }\quad \substack{
                           \text{IsolationGate} \\[5pt]
                           \land\;\text{RewardGate} \\[5pt]
                           \land\;\text{LengthGate}
                         }\,, \\[20pt]
\phi_{t-1} & \text{otherwise.}
\end{cases}
\label{eq:cgtr}
\end{equation}
where $t_{\text{last}}$ is the step of the most recent teacher refresh.
The three conditions are defined as follows.

\paragraph{Isolation gate.}
The teacher is held frozen for at least $M_{\min}$ steps after the
last refresh:
\begin{equation}
t - t_{\text{last}} \ge M_{\min}.
\label{eq:isolation-gate}
\end{equation}
% This enforces \emph{parameter independence}
% (Section~\ref{sec:analysis-summary}): during the isolation window the
% teacher cannot absorb any student fluctuation, ensuring
% $\KL(\student\,\|\,\teacher)$ remains a meaningful corrective signal.
% The condition is a hard prerequisite; the two state-awareness gates
% below are evaluated only once it is met.
This directly enforces teacher isolation: during the isolation window,
the teacher cannot absorb student fluctuations.  By preventing immediate
tracking, the isolation period helps preserve \emph{temporal independence}
(Section~\ref{sec:analysis-summary}), keeping the distillation target
anchored to an earlier parameter state that can counter current student
deviations.
The condition is a hard prerequisite; the two state-aware gates
below are evaluated only once it is met.

\paragraph{Reward gate.}
The mean training reward over the most recent $w$ steps must exceed
the reward recorded at the last teacher refresh by at least $\delta_r$:
\begin{equation}
\bar{r}_{t-w:t} > r_{\mathrm{ref}} + \delta_r,
\label{eq:reward-gate}
\end{equation}
where $r_{\mathrm{ref}}$ is the rolling-average reward captured at the last refresh (updated after each successful refresh),
and $\delta_r \ge 0$ is the improvement threshold.
$\bar{r}_{t-w:t}$ is computed online from training rollouts and
requires no additional model inference.
This implements a \emph{ratchet}: $r_{\mathrm{ref}}$ rises with each
refresh, so every subsequent teacher update requires strictly higher
performance than the previous one.

\paragraph{Length-tail gate.}
The maximum-to-mean sequence-length ratio over the most recent $w$
steps must remain below a safety threshold:
\begin{equation}
\frac{\max(\texttt{seqlen}_{t-w:t})}
     {\operatorname{mean}(\texttt{seqlen}_{t-w:t})}
\le L_{\max}.
\label{eq:length-gate}
\end{equation}
% This addresses the \emph{behavioral stability} dimension of state
% awareness: a rising relative length tail is a leading indicator of generation
% collapse (Section~\ref{sec:analysis-fragility}) that reward
% improvement alone cannot detect in time.
This makes refreshes sensitive to \emph{behavioral instability}: a rising
relative length tail is an early warning of unstable training
(Section~\ref{sec:analysis-fragility}) that reward improvement alone
cannot detect in time.

Compared to a fixed-$M$ schedule, CGTR introduces three interpretable
hyperparameters ($M_{\min}$, $\delta_r$, $L_{\max}$), each with a
direct physical interpretation: $M_{\min}$ bounds the isolation period,
$\delta_r$ sets the performance consolidation threshold, and $L_{\max}$
limits admissible relative length dispersion.
These are set once from schedule-sweep diagnostics, in contrast to
the opaque integer $M$ of a fixed schedule.
%%%%%%%%%%%%%%%%%%%%%%%%%%%%%%%%%%%%%%%%%%%%%%%%%%%%%%%%%%%
\section{Results}
\label{sec:results}
%%%%%%%%%%%%%%%%%%%%%%%%%%%%%%%%%%%%%%%%%%%%%%%%%%%%%%%%%%%

We evaluate CGTR and fixed-schedule baselines in a 300-step long-horizon
comparison (Chemistry), run per-gate ablations to verify each condition's contribution, and test three methods on Biology, Physics,
and ToolUse to assess cross-task generalization.
Table~\ref{tab:phase3} summarizes the Chemistry long-horizon results;
Table~\ref{tab:ablation} reports the gate ablation;
Table~\ref{tab:generalization} reports the cross-task comparison.

\begin{table}[H]
\vspace{-0.8em}
\caption{%
Long-horizon comparison results (300 steps, no auxiliary recovery).
Test\,fin/peak = test mean@16 at end / best step.
Ent\,mn/fin = entropy mean / final.
Coll.\ follows Appendix~\ref{sec:collapse-detection} (--- = none observed);
\#Ref = teacher refreshes.
}
\centering
\small
\setlength{\tabcolsep}{3pt}
\begin{tabular}{lcccccc}
\toprule
\textbf{Sched.} &
  \textbf{Test\,fin} &
  \textbf{Test\,peak} &
  \textbf{Ent\,mn} &
  \textbf{Ent\,fin} &
  \textbf{Coll.} &
  \textbf{\#Ref} \\
\midrule
$M{=}50$ & 0.000 & 0.744 & 0.539 & 0.525 & 189 & 6 \\
EMA-04   & 0.713 & 0.798 & 1.539 & 3.015 & --- & --- \\
CGTR     & \textbf{0.806} & \textbf{0.809} & 0.654 & 0.780 & --- & \textbf{3} \\
\bottomrule
\end{tabular}
\vspace{-1mm}
\label{tab:phase3}
\vspace{-0.3em}
\end{table}

\subsection{Experimental Setup}
\label{sec:exp-setup}

We evaluate base Qwen3-8B~\citep{yang2025qwen3} without instruction tuning
on the level-L3 Chemistry, Biology, and Physics subsets of
SciKnowEval~\citep{feng2024sciknoweval} and on ToolAlpaca
ToolUse~\citep{tang2023toolalpaca}, using exact-match binary rewards.
Chemistry uses a 1,890/210 train/test split. Training uses $4\times$A100
GPUs, FSDP, global batch size~16 ($n{=}4$ rollouts per question), AdamW
lr~$10^{-5}$, distillation top-$k{=}20$, and JSD $\beta{=}0.5$.
The primary metric is test mean@16; we also report bootstrap-estimated B@4.
The 100-step sweep compares $M\in\{1,2,3,4,5,8,10,20,50\}$, a fixed
teacher, and EMA $\alpha\in\{0.1,0.04\}$; Chemistry long-horizon and
cross-task runs use 300-step training on held-out data.
Appendix~\ref{sec:appendix-repro} provides full details.
% \paragraph{Datasets.}
% We evaluate on two benchmark families spanning four tasks:
% \begin{itemize}
%   \item \textbf{Science Q\&A} (Chemistry, Physics, Biology):
%     Undergraduate-level scientific multiple-choice reasoning using the
%     reasoning subsets (level~L3) from
%     SciKnowEval~\citep{feng2024sciknoweval}.
%     Each question provides four options; correctness is judged by exact
%     option match against the gold label.
%   \item \textbf{Tool use}: Mapping a tool-API specification and user
%     request to the correct tool call, using
%     ToolAlpaca~\citep{tang2023toolalpaca}.
%     The reward is a binary signal indicating whether the predicted tool
%     name and arguments exactly match the reference call.
% \end{itemize}

% We use Qwen3-8B~\citep{yang2025qwen3} (pretrained, no instruction-tuning)
% on SciKnowEval~\citep{feng2024sciknoweval} Chemistry (1,890/210 split).
% Training uses $4\times$A100 GPUs, FSDP, global batch size~16
% ($n{=}4$ rollouts per question), AdamW lr~$10^{-5}$, distillation
% top-$k{=}20$, JSD $\alpha{=}0.5$.
% The primary metric is $\text{test\_mean@16}$ (average correctness over 16
% sampled responses per test question); we also report bootstrap-estimated B@4.
% The schedule sweep covers $M \in \{1,2,3,4,5,8,10,20,50\}$, fixed
% teacher, and EMA $\alpha \in \{0.1, 0.04\}$, each evaluated at step~100.
% Long-horizon comparison (Chemistry) and cross-task runs (Biology, Physics,
% ToolUse) each use 300-step training on held-out data.
% Full experimental details are given in Appendix~\ref{sec:appendix-repro}.

\subsection{Fixed-Interval Refresh Collapses Under Long-Horizon Training}
\label{sec:results-fixedcollapse}

\begin{figure*}[t]
\centering
\includegraphics[width=\linewidth]{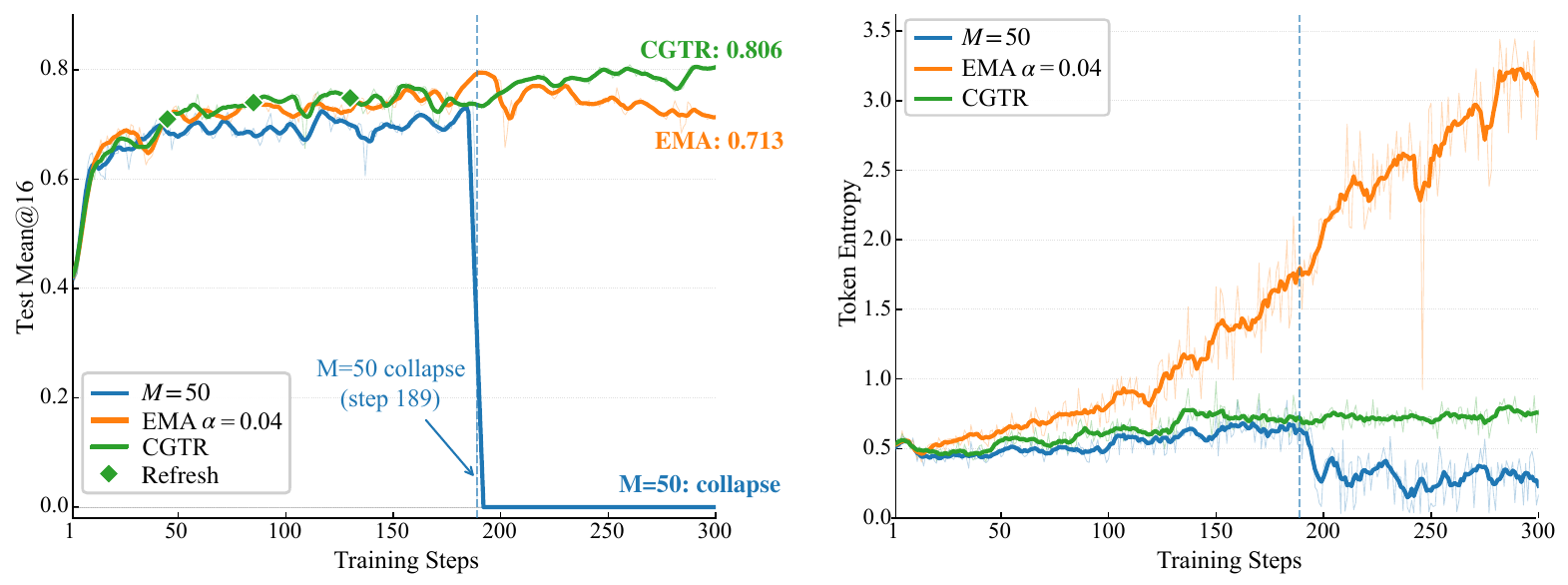}
\vspace{-6mm}
\caption{%
Long-horizon comparison (300 steps, Chemistry, no auxiliary recovery).
\textbf{(Left)}~Test mean@16: $M{=}50$ (blue) collapses at step~189;
EMA $\alpha{=}0.04$ (orange) plateaus at $0.713$;
CGTR (green) achieves $0.806$ with zero collapse and 3 adaptive refreshes.
\textbf{(Right)}~Token entropy: EMA entropy drifts upward throughout training,
while CGTR entropy remains stably low; $M{=}50$ exhibits a sharp entropy decrease after collapse.
}
\label{fig:300step}
\vspace{-0.8em}
\end{figure*}

Table~\ref{tab:phase3} and Figure~\ref{fig:300step} report the
long-horizon comparison; $M{=}50$ collapses catastrophically at step~189: $\text{mean@}16$
drops from its peak of 0.744 (step~187) to 0.000 and does not
recover for the remainder of training.
The mechanism is consistent with the structural prediction of
Section~\ref{sec:analysis-fragility}: the hard refresh at step~150
fires while the student is in an elevated entropy state ($\approx 0.98$),
copying a transiently unstable snapshot into the teacher.
By step~189, entropy collapses toward zero and the model degenerates
into repetitive outputs. Subsequent clock-driven refreshes copy the
degraded student state into the teacher, leaving no recovery path.

EMA $\alpha{=}0.04$ avoids catastrophic failure by capping per-step
contamination, but the trade-off is chronic drift: entropy grows
monotonically from $\approx 0.7$ to 3.0 by step~300, and performance
plateaus at 0.713 after step~190 with no further gain.

% These two failure modes---state-oblivious collapse ($M{=}50$) and
% steady-state contamination (EMA)---confirm that \emph{no fixed schedule
% can simultaneously provide isolation and resilience}: the very property
% that makes $M{=}50$ effective at short horizons is what renders it
% catastrophically fragile under extended training.

\subsection{CGTR Achieves Best Long-Horizon Performance Without Per-Dataset Tuning}
\label{sec:cgtr-results}

CGTR achieves the highest final score ($0.806$) and peak score ($0.809$)
in the Chemistry long-horizon run (Table~\ref{tab:phase3}), with zero
collapse over 300 steps.
Its entropy remains at a lower level on average than EMA's
($0.654$ vs.\ $1.539$).
The reward ratchet and length-tail guard ensure every
teacher update captures consolidation rather than transient drift,
avoiding failure modes that trap both fixed schedules.

\paragraph{Frozen-teacher controls.}
To test whether CGTR's gain comes simply from freezing the teacher,
we compare it with a teacher fixed at initialization and with the
$M{=}50$ teacher snapshot at step~100 frozen through step~300.
Both controls avoid collapse but finish below CGTR
($0.725$ and $0.765$ vs.\ $0.806$; Appendix~\ref{sec:appendix-fixed-controls}).
Thus, the gain is not reproduced by either simple freezing strategy
in this run; this comparison does not cover every possible freeze time.

% \paragraph{Gate mechanism: ratchet reward and length-tail guard.}
% The reward gate implements an implicit \emph{ratchet}: after each refresh
% the reference threshold is updated to the current reward level, so the
% next refresh requires strictly higher performance than the previous one.
% This ensures the teacher always represents a consolidated, improving state
% rather than a transiently elevated one.
% The length-tail gate provides a complementary safety check: even when
% the reward criterion is satisfied, an abnormally long generation tail
% ($L_{\text{tail}} > L_{\max}$) signals length-expansion instability and
% blocks the refresh until the student stabilizes.
% Together, these two gates encode the condition that a refresh is warranted
% only when performance has genuinely advanced \emph{and} the model is
% behaviorally stable---the two properties that a fixed schedule ignores.

\paragraph{Adaptive refresh behavior.}
CGTR triggers three teacher refreshes at steps~45, 85, and~130.
After the third refresh at step~130 (reward $= 0.844$), the threshold
rises to 0.894.  The student does not reach this level in the
remaining 170 steps, so the teacher stays frozen from step~130 to 300.
During this self-determined isolation window, test accuracy improves
from 0.752 to 0.806 ($+0.054$)---demonstrating that a long static
teacher is beneficial once the reward ratchet has ensured it encodes a
genuinely consolidated state.
% This contrasts sharply with $M{=}50$, whose refresh at step~150 also
% produces a 170-step freeze but without any state-readiness check,
% ultimately triggering collapse.

\subsection{Ablation: Each Gate Is Necessary}
\label{sec:ablation}

We remove one gate at a time on Chemistry (300 steps) to quantify
each condition's independent contribution (Table~\ref{tab:ablation}).

\begin{table}[t]
\caption{%
Gate ablation on Chemistry (300 steps).
Test\,fin/peak = test mean@16 at end / best step.
\#Ref = teacher refreshes; Coll.\ follows
Appendix~\ref{sec:collapse-detection} (--- = none observed).
}
\centering
\small
\setlength{\tabcolsep}{4pt}
\begin{tabular}{lcccc}
\toprule
\textbf{Variant} & \textbf{Test\,fin} & \textbf{Test\,peak} &
  \textbf{\#Ref} & \textbf{Coll.} \\
\midrule
CGTR (full)          & 0.806 & 0.809 & 3  & ---  \\
w/o Isolation gate   & 0.731 & 0.768 & 12 & ---  \\
w/o Reward gate      & 0.000 & 0.781 & 6  & 218  \\
w/o Length-tail gate & 0.771 & 0.792 & 4  & ---  \\
\bottomrule
\end{tabular}
\label{tab:ablation}
\vspace{-0.3em}
\end{table}

Removing the \textbf{reward gate} is the most damaging: without the
ratchet condition, refreshes fire during transient reward dips, doubling
the refresh count (6 vs.\ 3) and collapsing at step~218
(test\,fin${}=0.000$); the length-tail gate guards only against
generation explosion---it monitors seqlen tail growth, not reward-level
bad refreshes---and therefore cannot intercept an ill-timed copy
triggered during a reward dip before length pathology has emerged.
Removing the \textbf{isolation gate} ($M_{\min}{=}0$) quadruples the
refresh rate (12 vs.\ 3); without a structural freeze floor the teacher
tracks the student too closely, and chronic contamination during
reward-plateau windows degrades the final score to 0.731
without triggering hard collapse.
Removing the \textbf{length-tail gate} allows one premature refresh at
step~127 (reward above threshold but tail ratio $3.41{>}L_{\max}$),
slightly corrupting the teacher and dropping the final score by 0.035;
the run avoids collapse, confirming the length-tail gate as a
complementary guard against ill-timed updates.
Importantly, all ablated variants reach comparable peak scores
(0.768--0.792), yet their final outcomes diverge sharply
(0.000--0.771).  This shows that peak performance alone can obscure
schedule fragility, making long-horizon retention essential for evaluation.

\begin{table}[t]
\vspace{-0.4em}
\caption{%
Cross-task generalization (300 steps).
Test\,fin = final mean@16; Test\,peak = best step value.
Coll.\ follows Appendix~\ref{sec:collapse-detection} (--- = none observed).
}
\centering
\small
\setlength{\tabcolsep}{4pt}
\begin{tabular}{llcccc}
\toprule
\textbf{Task} & \textbf{Sched.} &
  \textbf{Test\,fin} & \textbf{Test\,peak} &
  \textbf{Ent\,fin} & \textbf{Coll.} \\
\midrule
\multirow{3}{*}{Biology}
  & $M{=}50$ & 0.000 & 0.560 & 0.000 & 130 \\
  & EMA-04   & 0.471 & 0.569 & 3.522 & --- \\
  & CGTR     & \textbf{0.485} & \textbf{0.573} & 0.204 & --- \\
\midrule
\multirow{3}{*}{Physics}
  & $M{=}50$ & 0.670 & 0.702 & 0.450 & --- \\
  & EMA-04   & 0.660 & 0.728 & 2.271 & --- \\
  & CGTR     & \textbf{0.703} & \textbf{0.735} & 0.278 & --- \\
\midrule
\multirow{3}{*}{ToolUse}
  & $M{=}50$ & 0.467 & 0.612 & 1.843 & --- \\
  & EMA-04   & 0.481 & 0.641 & 3.206 & --- \\
  & CGTR     & \textbf{0.536} & \textbf{0.643} & \textbf{0.892} & --- \\
\bottomrule
\end{tabular}
\label{tab:generalization}
\vspace{-0.8em}
\end{table}

\paragraph{Cross-task generalization.}
We run all three schedules for 300 steps on Biology, Physics, and ToolUse
to test whether the failure modes generalize.
The cross-task results (Table~\ref{tab:generalization}) are consistent
across all four settings.
\textbf{Biology:} $M{=}50$ collapses at step~130 (final $\to 0.000$),
replicating the Chemistry failure on a different task; CGTR avoids
collapse and outperforms EMA (final 0.485 vs.\ 0.471), triggering twice
(steps~127, 192).
\textbf{Physics:} all three schedules avoid collapse; CGTR still leads
on final test (0.703), peak (0.735), and entropy (0.278), with a single
gate trigger at step~180 reflecting Physics' characteristically slower reward dynamics.
\textbf{ToolUse:} no hard collapse; $M{=}50$ nonetheless declines from
peak 0.612 to final 0.467 under state-oblivious refreshes, EMA declines
similarly (final 0.481, entropy 3.206), while CGTR holds the best final
(0.536) and lowest entropy (0.892) with two well-timed refreshes (steps~95, 190).

\paragraph{Multi-seed verification.}
Across three seeds on each of Chemistry and Biology, $M{=}50$ collapses
in all runs (3/3 per task), whereas CGTR has no observed collapse
(0/3 per task).
CGTR's final mean@16 is $0.805\pm0.007$ on Chemistry and
$0.482\pm0.003$ on Biology (mean$\pm$std;
Appendix~\ref{sec:appendix-multiseed}).

% \paragraph{Cross-model verification.}
% Single-seed Chemistry runs on Qwen3-1.7B and Llama-3.1-8B preserve
% the ordering CGTR $>$ EMA-04 $>$ $M{=}50$ under the shared CGTR
% thresholds. CGTR finishes at $0.652$ and $0.723$, respectively,
% while $M{=}50$ collapses in both runs
% (Appendix~\ref{sec:appendix-cross-model}).

\textbf{Summary.}
Across all four tasks, CGTR achieves the best final and peak scores with
no observed collapse, using one shared parameter set without per-dataset
retuning.  All thresholds were fixed from the 100-step Chemistry sweep
before the long-horizon and cross-task evaluations; selection and
sensitivity are documented in Appendices~\ref{sec:appendix-cgtr-hparams}
and~\ref{sec:appendix-sensitivity}.
% CGTR achieves the best final and peak score on all four tasks, along
% with a \textbf{zero-collapse rate}, while $M{=}50$ collapses in two
% of four (Chemistry at step~189, Biology at step~130).
% CGTR resolves this with \textbf{a single shared parameter set}: its gate
% conditions self-regulate refresh frequency to each task's distinct dynamics---3 refreshes in Chemistry, 2 in Biology, 1 in Physics, and
% 2 in ToolUse---without any per-dataset retuning.
% Critically, all gate thresholds were fixed prior to the long-horizon
% and cross-task experiments, based solely on the 100-step schedule-sweep
% diagnostics on Chemistry; the Biology, Physics, and ToolUse results therefore
% constitute a clean held-out evaluation of the chosen parameters across tasks (see Appendix~\ref{sec:appendix-cgtr-hparams} for selection rationale
% and Appendix~\ref{sec:appendix-sensitivity} for a sensitivity analysis confirming that performance is stable and collapse-free across a range of threshold values).

%%%%%%%%%%%%%%%%%%%%%%%%%%%%%%%%%%%%%%%%%%%%%%%%%%%%%%%%%%%
\section{Conclusion}
\label{sec:conclusion}
%%%%%%%%%%%%%%%%%%%%%%%%%%%%%%%%%%%%%%%%%%%%%%%%%%%%%%%%%%%

% We have studied temporal coupling in self on-policy distillation and
% shown that teacher isolation periods---not teacher age---are the key
% mechanism enabling stable learning.
% In our setting, sparse hard refresh ($M{=}50$) achieves the best
% short-horizon performance precisely because complete teacher freezing
% prevents per-step contamination; however, this same state-oblivious
% schedule collapses catastrophically under long-horizon training
% ($\text{mean@}16 \to 0.000$ at step~189) when the refresh fires during
% a period of student drift.
% We propose \emph{Consolidation-Gated Teacher Refresh} (CGTR), which
% enforces a minimum isolation window and gates each refresh on joint
% evidence of reward improvement and length-tail safety.
% CGTR achieves zero collapse and the highest final score ($0.806$) on
% Chemistry, and generalizes across four tasks (Chemistry, Biology,
% Physics, ToolUse) using a single shared parameter set without
% per-task retuning.
% The key design principle is: \emph{the question is not ``how old should
% the teacher be?'' but ``is the student ready for the teacher to move?''}

In this work, we show that \emph{isolation periods}---not teacher age---are the
structural property that enables stable self on-policy distillation, and
that fixed-interval schedules carrying this advantage nonetheless fail
catastrophically at long horizons through \emph{state-oblivious
collapse}: a single ill-timed refresh irreversibly copies a transiently
drifting student into the teacher, a failure mode mechanistically
distinct from EMA's chronic contamination and invisible under
short-horizon evaluation.
Our proposed \emph{Consolidation-Gated Teacher Refresh} (CGTR) resolves
this tension by preserving isolation periods while gating each refresh
on joint evidence of reward improvement and length-tail safety, ensuring
every teacher movement is a response to genuine student consolidation
rather than a clock signal.
CGTR achieves zero collapse and the best final score on all four tasks
(Chemistry, Biology, Physics, ToolUse) with a single shared parameter
set---demonstrating that the right question for teacher scheduling is
not \emph{``how old should the teacher be?''} but
\emph{``is the student ready for the teacher to move?''}
% %%%%%%%%%%%%%%%%%%%%%%%%%%%%%%%%%%%%%%%%%%%%%%%%%%%%%%%%%%%
% \section*{Limitations}
% \label{sec:limitations}
% %%%%%%%%%%%%%%%%%%%%%%%%%%%%%%%%%%%%%%%%%%%%%%%%%%%%%%%%%%%

% Our main experiments use Qwen3-8B across four tasks at moderate
% training scale. Additional single-seed Chemistry runs cover Qwen3-1.7B
% and Llama-3.1-8B.
% Extending the multi-seed analysis to additional architectures, larger
% models, longer horizons, and a broader range of task types remains an
% open direction.
% The reward gate currently relies on a binary correctness signal;
% whether denser or learned reward signals interact differently with
% the gating dynamics is an open question.

\subsection*{AI use statement}

Generative AI tools were used to improve the readability and presentation
of author-written text, including language polishing and LaTeX formatting,
and to support literature retrieval and discovery. The authors manually
reviewed every AI-assisted edit and verified all identified references
against the original publications. These tools were not used to generate
research ideas or hypotheses, design or execute the methodology or
experiments, implement the proposed method, generate or clean datasets,
conduct data analysis, interpret results, formulate or prove mathematical
claims, translate text, or draft manuscript sections. The authors take
full responsibility for the final content of this work.

% This work focuses on a specific and well-controlled setting; we
% outline directions for future extension.

% \paragraph{Broader model and task coverage.}
% Our experiments use Qwen3-8B on SciKnowEval (Chemistry, Biology,
% Physics) and a ToolUse benchmark.  The core phenomena---reference
% collapse under tight coupling, isolation-period advantage, and
% state-oblivious collapse---are consistently observed across all four
% tasks, suggesting they reflect general properties of self on-policy
% distillation rather than dataset-specific artifacts.  Extending the
% analysis to other model families and task domains is a natural next step.

% \paragraph{Theoretical characterization.}
% Our study is empirical and diagnostic.  A formal account of when
% consolidation gating is sufficient to prevent long-horizon collapse, and
% how the gate thresholds interact with the underlying learning dynamics,
% would complement the empirical findings and is left for future work.

% \paragraph{Adaptive gate design.}
% CGTR's gate parameters ($M_{\min}$, $\delta_r$, $L_{\max}$) are fixed
% across tasks and carry direct physical interpretations tied to observed
% training dynamics.
% Exploring learned or online-adaptive variants of the gating rule
% is an interesting direction that we leave open.

% Bibliography
\bibliographystyle{iclr2027_conference}
\bibliography{custom}

\begin{thebibliography}{22}
\providecommand{\natexlab}[1]{#1}
\providecommand{\url}[1]{\texttt{#1}}
\expandafter\ifx\csname urlstyle\endcsname\relax
  \providecommand{\doi}[1]{doi: #1}\else
  \providecommand{\doi}{doi: \begingroup \urlstyle{rm}\Url}\fi

\bibitem[Agarwal et~al.(2024)Agarwal, Vieillard, Zhou, Stanczyk, Ramos, Geist, and Bachem]{agarwal2024onpolicy}
Rishabh Agarwal, Nino Vieillard, Yongchao Zhou, Piotr Stanczyk, Sabela Ramos, Matthieu Geist, and Olivier Bachem.
\newblock On-policy distillation of language models: Learning from self-generated mistakes.
\newblock In \emph{International Conference on Learning Representations}, 2024.
\newblock \doi{10.48550/arXiv.2306.13649}.
\newblock URL \url{https://arxiv.org/abs/2306.13649}.
\newblock arXiv:2306.13649.

\bibitem[Feng et~al.(2025)Feng, Shen, Wang, Zhuang, Tang, Zhang, and Ding]{feng2024sciknoweval}
Kehua Feng, Xinyi Shen, Weijie Wang, Xiang Zhuang, Yuqi Tang, Qiang Zhang, and Keyan Ding.
\newblock Sciknoweval: Evaluating multi-level scientific knowledge of large language models, 2025.
\newblock URL \url{https://arxiv.org/abs/2406.09098}.

\bibitem[Grill et~al.(2020)Grill, Strub, Altch{\'e}, Tallec, Richemond, Buchatskaya, Doersch, {Avila Pires}, Guo, Azar, Piot, Kavukcuoglu, Munos, and Valko]{grill2020bootstrap}
Jean-Bastien Grill, Florian Strub, Florent Altch{\'e}, Corentin Tallec, Pierre~H. Richemond, Elena Buchatskaya, Carl Doersch, Bernardo {Avila Pires}, Zhaohan~Daniel Guo, Mohammad~Gheshlaghi Azar, Bilal Piot, Koray Kavukcuoglu, R{\'e}mi Munos, and Michal Valko.
\newblock Bootstrap your own latent: A new approach to self-supervised learning.
\newblock In \emph{Advances in Neural Information Processing Systems}, volume~33, pp.\  21271--21284, 2020.
\newblock URL \url{https://arxiv.org/abs/2006.07733}.

\bibitem[Guo et~al.(2025)Guo, Yang, Zhang, Song, Wang, Zhu, Xu, Zhang, Ma, Bi, Zhang, Yu, Wu, Wu, Gou, Shao, Li, Gao, Liu, Xue, Wang, Wu, Feng, Lu, Zhao, Deng, Ruan, Dai, Chen, Ji, Li, Lin, Dai, Luo, Hao, Chen, Li, Zhang, Xu, Ding, Gao, Qu, Li, Guo, Li, Chen, Yuan, Tu, Qiu, Li, Cai, Ni, Liang, Chen, Dong, Hu, You, Gao, Guan, Huang, Yu, Wang, Zhang, Zhao, Wang, Zhang, Xu, Xia, Zhang, Zhang, Tang, Zhou, Li, Wang, Li, Tian, Huang, Zhang, Wang, Chen, Du, Ge, Zhang, Pan, Wang, Chen, Jin, Chen, Lu, Zhou, Chen, Ye, Wang, Yu, Zhou, Pan, Li, Zhou, Wu, Yun, Pei, Sun, Wang, Zeng, Liu, Liang, Gao, Yu, Zhang, Xiao, An, Liu, Wang, Chen, Nie, Cheng, Liu, Xie, Liu, Yang, Li, Su, Lin, Li, Jin, Shen, Chen, Sun, Wang, Song, Zhou, Wang, Shan, Li, Wang, Wei, Zhang, Xu, Li, Zhao, Sun, Wang, Yu, Zhang, Shi, Xiong, He, Piao, Wang, Tan, Ma, Liu, Guo, Ou, Wang, Gong, Zou, He, Xiong, Luo, You, Liu, Zhou, Zhu, Huang, Li, Zheng, Zhu, Ma, Tang, Zha, Yan, Ren, Ren, Sha, Fu, Xu, Xie, Zhang, Hao, Ma, Yan, Wu, Gu, Zhu, Liu, Li, Xie, Song,
  Pan, Huang, Xu, Zhang, and Zhang]{guo2025deepseekr1}
Daya Guo, Dejian Yang, Haowei Zhang, Junxiao Song, Peiyi Wang, Qihao Zhu, Runxin Xu, Ruoyu Zhang, Shirong Ma, Xiao Bi, Xiaokang Zhang, Xingkai Yu, Yu~Wu, Z.~F. Wu, Zhibin Gou, Zhihong Shao, Zhuoshu Li, Ziyi Gao, Aixin Liu, Bing Xue, Bingxuan Wang, Bochao Wu, Bei Feng, Chengda Lu, Chenggang Zhao, Chengqi Deng, Chong Ruan, Damai Dai, Deli Chen, Dongjie Ji, Erhang Li, Fangyun Lin, Fucong Dai, Fuli Luo, Guangbo Hao, Guanting Chen, Guowei Li, H.~Zhang, Hanwei Xu, Honghui Ding, Huazuo Gao, Hui Qu, Hui Li, Jianzhong Guo, Jiashi Li, Jingchang Chen, Jingyang Yuan, Jinhao Tu, Junjie Qiu, Junlong Li, J.~L. Cai, Jiaqi Ni, Jian Liang, Jin Chen, Kai Dong, Kai Hu, Kaichao You, Kaige Gao, Kang Guan, Kexin Huang, Kuai Yu, Lean Wang, Lecong Zhang, Liang Zhao, Litong Wang, Liyue Zhang, Lei Xu, Leyi Xia, Mingchuan Zhang, Minghua Zhang, Minghui Tang, Mingxu Zhou, Meng Li, Miaojun Wang, Mingming Li, Ning Tian, Panpan Huang, Peng Zhang, Qiancheng Wang, Qinyu Chen, Qiushi Du, Ruiqi Ge, Ruisong Zhang, Ruizhe Pan, Runji Wang, R.~J.
  Chen, R.~L. Jin, Ruyi Chen, Shanghao Lu, Shangyan Zhou, Shanhuang Chen, Shengfeng Ye, Shiyu Wang, Shuiping Yu, Shunfeng Zhou, Shuting Pan, S.~S. Li, Shuang Zhou, Shaoqing Wu, Tao Yun, Tian Pei, Tianyu Sun, T.~Wang, Wangding Zeng, Wen Liu, Wenfeng Liang, Wenjun Gao, Wenqin Yu, Wentao Zhang, W.~L. Xiao, Wei An, Xiaodong Liu, Xiaohan Wang, Xiaokang Chen, Xiaotao Nie, Xin Cheng, Xin Liu, Xin Xie, Xingchao Liu, Xinyu Yang, Xinyuan Li, Xuecheng Su, Xuheng Lin, X.~Q. Li, Xiangyue Jin, Xiaojin Shen, Xiaosha Chen, Xiaowen Sun, Xiaoxiang Wang, Xinnan Song, Xinyi Zhou, Xianzu Wang, Xinxia Shan, Y.~K. Li, Y.~Q. Wang, Y.~X. Wei, Yang Zhang, Yanhong Xu, Yao Li, Yao Zhao, Yaofeng Sun, Yaohui Wang, Yi~Yu, Yichao Zhang, Yifan Shi, Yiliang Xiong, Ying He, Yishi Piao, Yisong Wang, Yixuan Tan, Yiyang Ma, Yiyuan Liu, Yongqiang Guo, Yuan Ou, Yuduan Wang, Yue Gong, Yuheng Zou, Yujia He, Yunfan Xiong, Yuxiang Luo, Yuxiang You, Yuxuan Liu, Yuyang Zhou, Y.~X. Zhu, Yanping Huang, Yaohui Li, Yi~Zheng, Yuchen Zhu, Yunxian Ma, Ying
  Tang, Yukun Zha, Yuting Yan, Z.~Z. Ren, Zehui Ren, Zhangli Sha, Zhe Fu, Zhean Xu, Zhenda Xie, Zhengyan Zhang, Zhewen Hao, Zhicheng Ma, Zhigang Yan, Zhiyu Wu, Zihui Gu, Zijia Zhu, Zijun Liu, Zilin Li, Ziwei Xie, Ziyang Song, Zizheng Pan, Zhen Huang, Zhipeng Xu, Zhongyu Zhang, and Zhen Zhang.
\newblock Deepseek-r1 incentivizes reasoning in llms through reinforcement learning.
\newblock \emph{Nature}, 645\penalty0 (8081):\penalty0 633–638, September 2025.
\newblock ISSN 1476-4687.
\newblock \doi{10.1038/s41586-025-09422-z}.
\newblock URL \url{http://dx.doi.org/10.1038/s41586-025-09422-z}.

\bibitem[Hinton et~al.(2015)Hinton, Vinyals, and Dean]{hinton2015distilling}
Geoffrey Hinton, Oriol Vinyals, and Jeff Dean.
\newblock Distilling the knowledge in a neural network.
\newblock \emph{arXiv preprint arXiv:1503.02531}, 2015.
\newblock URL \url{https://arxiv.org/abs/1503.02531}.

\bibitem[H{\"u}botter et~al.(2026)H{\"u}botter, L{\"u}beck, Behric, Baumann, Bagatella, Marta, Hakimi, Shenfeld, Kleine~Buening, Guestrin, and Krause]{hubotter2026reinforcement}
Jonas H{\"u}botter, Frederike L{\"u}beck, Lejs Behric, Anton Baumann, Marco Bagatella, Daniel Marta, Ido Hakimi, Idan Shenfeld, Thomas Kleine~Buening, Carlos Guestrin, and Andreas Krause.
\newblock Reinforcement learning via self-distillation, 2026.
\newblock URL \url{https://arxiv.org/abs/2601.20802}.

\bibitem[Kim \& Rush(2016)Kim and Rush]{kim-rush-2016-sequence}
Yoon Kim and Alexander~M. Rush.
\newblock Sequence-level knowledge distillation.
\newblock In \emph{Proceedings of the 2016 Conference on Empirical Methods in Natural Language Processing}, pp.\  1317--1327. Association for Computational Linguistics, 2016.
\newblock \doi{10.18653/v1/D16-1139}.

\bibitem[Li et~al.(2026)Li, Zuo, He, Zhang, Xiao, Qian, Yu, Gao, Yang, Liu, and Ding]{li2026rethinking}
Yaxuan Li, Yuxin Zuo, Bingxiang He, Jinqian Zhang, Chaojun Xiao, Cheng Qian, Tianyu Yu, Huan-ang Gao, Wenkai Yang, Zhiyuan Liu, and Ning Ding.
\newblock Rethinking on-policy distillation of large language models: Phenomenology, mechanism, and recipe, 2026.
\newblock URL \url{https://arxiv.org/abs/2604.13016}.

\bibitem[Loshchilov \& Hutter(2019)Loshchilov and Hutter]{loshchilov2018decoupled}
Ilya Loshchilov and Frank Hutter.
\newblock Decoupled weight decay regularization.
\newblock In \emph{International Conference on Learning Representations}, 2019.
\newblock URL \url{https://arxiv.org/abs/1711.05101}.

\bibitem[Mnih et~al.(2015)Mnih, Kavukcuoglu, Silver, Rusu, Veness, Bellemare, Graves, Riedmiller, Fidjeland, Ostrovski, Petersen, Beattie, Sadik, Antonoglou, King, Kumaran, Wierstra, Legg, and Hassabis]{mnih2015human}
Volodymyr Mnih, Koray Kavukcuoglu, David Silver, Andrei~A. Rusu, Joel Veness, Marc~G. Bellemare, Alex Graves, Martin Riedmiller, Andreas~K. Fidjeland, Georg Ostrovski, Stig Petersen, Charles Beattie, Amir Sadik, Ioannis Antonoglou, Helen King, Dharshan Kumaran, Daan Wierstra, Shane Legg, and Demis Hassabis.
\newblock Human-level control through deep reinforcement learning.
\newblock \emph{Nature}, 518\penalty0 (7540):\penalty0 529--533, 2015.
\newblock \doi{10.1038/nature14236}.

\bibitem[Ouyang et~al.(2022)Ouyang, Wu, Jiang, Almeida, Wainwright, Mishkin, Zhang, Agarwal, Slama, Ray, Schulman, Hilton, Kelton, Miller, Simens, Askell, Welinder, Christiano, Leike, and Lowe]{ouyang2022training}
Long Ouyang, Jeffrey Wu, Xu~Jiang, Diogo Almeida, Carroll~L. Wainwright, Pamela Mishkin, Chong Zhang, Sandhini Agarwal, Katarina Slama, Alex Ray, John Schulman, Jacob Hilton, Fraser Kelton, Luke Miller, Maddie Simens, Amanda Askell, Peter Welinder, Paul Christiano, Jan Leike, and Ryan Lowe.
\newblock Training language models to follow instructions with human feedback.
\newblock \emph{Advances in Neural Information Processing Systems}, 35:\penalty0 27730--27744, 2022.
\newblock URL \url{https://arxiv.org/abs/2203.02155}.

\bibitem[Polyak \& Juditsky(1992)Polyak and Juditsky]{polyak1992acceleration}
Boris~T. Polyak and Anatoli~B. Juditsky.
\newblock Acceleration of stochastic approximation by averaging.
\newblock \emph{{SIAM} Journal on Control and Optimization}, 30\penalty0 (4):\penalty0 838--855, 1992.
\newblock \doi{10.1137/0330046}.

\bibitem[Rusu et~al.(2015)Rusu, Colmenarejo, G{\"{u}}l{\c{c}}ehre, Desjardins, Kirkpatrick, Pascanu, Mnih, Kavukcuoglu, and Hadsell]{rusu2015policy}
Andrei~A. Rusu, Sergio~Gomez Colmenarejo, {\c{C}}aglar G{\"{u}}l{\c{c}}ehre, Guillaume Desjardins, James Kirkpatrick, Razvan Pascanu, Volodymyr Mnih, Koray Kavukcuoglu, and Raia Hadsell.
\newblock Policy distillation.
\newblock \emph{arXiv preprint arXiv:1511.06295}, 2015.
\newblock URL \url{https://arxiv.org/abs/1511.06295}.

\bibitem[Sang et~al.(2026)Sang, Xu, Zhou, He, Wang, and Sun]{sang2026crisp}
Hejian Sang, Yuanda Xu, Zhengze Zhou, Ran He, Zhipeng Wang, and Jiachen Sun.
\newblock Crisp: Compressed reasoning via iterative self-policy distillation, 2026.
\newblock URL \url{https://arxiv.org/abs/2603.05433}.

\bibitem[Schulman et~al.(2017)Schulman, Wolski, Dhariwal, Radford, and Klimov]{schulman2017proximal}
John Schulman, Filip Wolski, Prafulla Dhariwal, Alec Radford, and Oleg Klimov.
\newblock Proximal policy optimization algorithms.
\newblock \emph{arXiv preprint arXiv:1707.06347}, 2017.
\newblock URL \url{https://arxiv.org/abs/1707.06347}.

\bibitem[Shao et~al.(2024)Shao, Wang, Zhu, Xu, Song, Bi, Zhang, Zhang, Li, Wu, and Guo]{shao2024deepseekmath}
Zhihong Shao, Peiyi Wang, Qihao Zhu, Runxin Xu, Junxiao Song, Xiao Bi, Haowei Zhang, Mingchuan Zhang, Y.~K. Li, Y.~Wu, and Daya Guo.
\newblock Deepseekmath: Pushing the limits of mathematical reasoning in open language models, 2024.
\newblock URL \url{https://arxiv.org/abs/2402.03300}.

\bibitem[Tang et~al.(2023)Tang, Deng, Lin, Han, Liang, Cao, and Sun]{tang2023toolalpaca}
Qiaoyu Tang, Ziliang Deng, Hongyu Lin, Xianpei Han, Qiao Liang, Boxi Cao, and Le~Sun.
\newblock Toolalpaca: Generalized tool learning for language models with 3000 simulated cases, 2023.
\newblock URL \url{https://arxiv.org/abs/2306.05301}.

\bibitem[Tarvainen \& Valpola(2017)Tarvainen and Valpola]{tarvainen2017mean}
Antti Tarvainen and Harri Valpola.
\newblock Mean teachers are better role models: Weight-averaged consistency targets improve semi-supervised deep learning results.
\newblock In \emph{Advances in Neural Information Processing Systems}, volume~30, 2017.

\bibitem[Yang et~al.(2025)Yang, Li, Yang, Zhang, Hui, Zheng, Yu, Gao, Huang, Lv, Zheng, Liu, Zhou, Huang, Hu, Ge, Wei, Lin, Tang, Yang, Tu, Zhang, Yang, Yang, Zhou, Zhou, Lin, Dang, Bao, Yang, Yu, Deng, Li, Xue, Li, Zhang, Wang, Zhu, Men, Gao, Liu, Luo, Li, Tang, Yin, Ren, Wang, Zhang, Ren, Fan, Su, Zhang, Zhang, Wan, Liu, Wang, Cui, Zhang, Zhou, and Qiu]{yang2025qwen3}
An~Yang, Anfeng Li, Baosong Yang, Beichen Zhang, Binyuan Hui, Bo~Zheng, Bowen Yu, Chang Gao, Chengen Huang, Chenxu Lv, Chujie Zheng, Dayiheng Liu, Fan Zhou, Fei Huang, Feng Hu, Hao Ge, Haoran Wei, Huan Lin, Jialong Tang, Jian Yang, Jianhong Tu, Jianwei Zhang, Jianxin Yang, Jiaxi Yang, Jing Zhou, Jingren Zhou, Junyang Lin, Kai Dang, Keqin Bao, Kexin Yang, Le~Yu, Lianghao Deng, Mei Li, Mingfeng Xue, Mingze Li, Pei Zhang, Peng Wang, Qin Zhu, Rui Men, Ruize Gao, Shixuan Liu, Shuang Luo, Tianhao Li, Tianyi Tang, Wenbiao Yin, Xingzhang Ren, Xinyu Wang, Xinyu Zhang, Xuancheng Ren, Yang Fan, Yang Su, Yichang Zhang, Yinger Zhang, Yu~Wan, Yuqiong Liu, Zekun Wang, Zeyu Cui, Zhenru Zhang, Zhipeng Zhou, and Zihan Qiu.
\newblock Qwen3 technical report.
\newblock \emph{arXiv preprint arXiv:2505.09388}, 2025.
\newblock URL \url{https://arxiv.org/abs/2505.09388}.

\bibitem[Yang et~al.(2026)Yang, Qin, Si, Chen, Gu, Yao, Lin, Wang, Wang, and Duan]{yang2026selfdistilledrlvr}
Chenxu Yang, Chuanyu Qin, Qingyi Si, Minghui Chen, Naibin Gu, Dingyu Yao, Zheng Lin, Weiping Wang, Jiaqi Wang, and Nan Duan.
\newblock Self-distilled rlvr, 2026.
\newblock URL \url{https://arxiv.org/abs/2604.03128}.

\bibitem[Zelikman et~al.(2022)Zelikman, Wu, Mu, and Goodman]{zelikman2022star}
Eric Zelikman, Yuhuai Wu, Jesse Mu, and Noah Goodman.
\newblock Star: Bootstrapping reasoning with reasoning.
\newblock In S.~Koyejo, S.~Mohamed, A.~Agarwal, D.~Belgrave, K.~Cho, and A.~Oh (eds.), \emph{Advances in Neural Information Processing Systems}, volume~35, pp.\  15476--15488. Curran Associates, Inc., 2022.
\newblock \doi{10.52202/068431-1126}.
\newblock URL \url{https://proceedings.neurips.cc/paper_files/paper/2022/file/639a9a172c044fbb64175b5fad42e9a5-Paper-Conference.pdf}.

\bibitem[Zhao et~al.(2026)Zhao, Xie, Liu, Huang, Pang, Chen, and Grover]{zhao2026selfdistilled}
Siyan Zhao, Zhihui Xie, Mengchen Liu, Jing Huang, Guan Pang, Feiyu Chen, and Aditya Grover.
\newblock Self-distilled reasoner: On-policy self-distillation for large language models, 2026.
\newblock URL \url{https://arxiv.org/abs/2601.18734}.

\end{thebibliography}

\newpage
\appendix

%%%%%%%%%%%%%%%%%%%%%%%%%%%%%%%%%%%%%%%%%%%%%%%%%%%%%%%%%%%
\section{Experimental Setup and Evaluation Protocol}
\label{sec:appendix-repro}
%%%%%%%%%%%%%%%%%%%%%%%%%%%%%%%%%%%%%%%%%%%%%%%%%%%%%%%%%%%

\subsection{Data and Prompt Format}

\paragraph{Datasets.}
We evaluate on two benchmark families spanning four tasks:
\begin{itemize}
  \item \textbf{Science Q\&A} (Chemistry, Physics, Biology):
    Undergraduate-level scientific multiple-choice reasoning using the
    reasoning subsets (level~L3) from
    SciKnowEval~\citep{feng2024sciknoweval}.
    Each question provides four options; correctness is judged by exact
    option match against the gold label.
  \item \textbf{Tool use}: Mapping a tool-API specification and user
    request to the correct tool call, using
    ToolAlpaca~\citep{tang2023toolalpaca}.
    The reward is a binary signal indicating whether the predicted tool
    name and arguments exactly match the reference call.
\end{itemize}

\paragraph{Dataset splits.}
SciKnowEval~\citep{feng2024sciknoweval} is split 9:1 into
train and test sets: Chemistry 1,890/210, Biology 450/50,
Physics 720/80.
ToolAlpaca~\citep{tang2023toolalpaca} contains 3,938 tool-use instances
covering 426 real-world APIs; we use a 9:1 split (3,544 train / 394 test).
No question appears in both splits; the test set is held out
throughout training.

\paragraph{Prompt format.}
\textbf{Science Q\&A.}
Each question is presented as a two-turn conversation.
The system turn contains a fixed instruction:
\begin{quote}\small
\textit{Given a question and four options, please select the right
answer.  Respond in the following format: \texttt{<reasoning>}
\ldots\texttt{</reasoning><answer>}\ldots\texttt{</answer>}.
For the answer, only output the letter (A, B, C, or D).}
\end{quote}
The user turn contains the question body with options appended.
\textbf{ToolUse.}
The prompt contains the tool-API specification and user request, and
the response is the predicted tool call.
Across tasks, the maximum prompt length is 2,048 tokens and the maximum
generated response length is 8,192 tokens.

\paragraph{Self-teacher input.}
Following SDPO~\citep{hubotter2026reinforcement}, the student receives
the original prompt $x$.  For each rollout $y$, the teacher re-evaluates
$y$ token by token under $x$ augmented with a successful response from
the same rollout group when available.  Otherwise, available environment
feedback is used; without either source, the sample provides no
self-distillation target.  A sample's own success is not reused.

\subsection{Reward Function}

\textbf{Science Q\&A.}
The reward is a binary correctness signal:
\[
  r = \mathbf{1}\bigl[\text{extract}(\hat{y}) = y^*\bigr],
\]
where $\hat{y}$ is the model response and $y^*$ is the
ground-truth option letter.
Extraction parses the last \texttt{<answer>}$\ldots$\texttt{</answer>}
tag in the response; if parsing fails the reward is 0.
\textbf{ToolUse.}
The reward is 1 only when both the predicted tool name and argument
string exactly match the reference call; otherwise it is 0.
No external verifier or neural reward model is used.
Ground-truth answers and reference calls are not inserted into student
prompts; teacher-only context follows the feedback construction above.

\subsection{Rollout and Decoding Parameters}

\paragraph{Training rollouts.}
Temperature~$1.0$, top-$p{=}1.0$, top-$k{=}{-}1$ (disabled),
$n{=}4$ responses per question (GRPO group size~\citep{shao2024deepseekmath}).

\paragraph{Test evaluation.}
Following SDPO~\citep{hubotter2026reinforcement}, we sample
$n{=}16$ responses per question using temperature~$0.6$,
top-$p{=}0.95$, top-$k{=}{-}1$ (disabled), and
\texttt{do\_sample=True}.
Mean@16 is the mean correctness across the 16 sampled responses,
averaged over test questions.
B@4 follows the SDPO estimator: for each question, we draw 1,000
size-4 samples with replacement from its 16 validation responses,
take the maximum reward in each sample, average over the bootstrap
samples, and then average over test questions.
Schedule sweep: one evaluation at step~100.
Long-horizon comparison (Chemistry): every step (\texttt{test\_freq}$=1$).
Cross-task runs (Biology, Physics, ToolUse): every 10 steps.

\subsection{Optimizer and Training Details}

AdamW~\citep{loshchilov2018decoupled}: $\beta_1{=}0.9$,
$\beta_2{=}0.999$, weight decay~$0.01$, gradient clipping
$\ell_2$-norm~$1.0$, lr~$10^{-5}$, constant schedule.
The long-horizon comparison uses a 10-step linear warm-up; the schedule sweep uses none.
PPO~\citep{schulman2017proximal} mini-batch size~4 with gradient accumulation.

\subsection{Distillation Loss}

The distillation objective is a token-level top-$k$ approximation
of the generalized Jensen--Shannon Divergence (JSD).
For each token position, the top-$k{=}20$ logit indices of the
\emph{student} are selected; teacher log-probabilities are
gathered at the same positions.
A tail bucket absorbs the residual probability mass
$1 - \sum_{i\in\text{top-}k}p_i$, computed via
\texttt{logsumexp} for numerical stability,
yielding a $(k{+}1)$-class categorical for both student and teacher.
Let $p_j{=}\pi_\theta(\cdot\mid x,y_{<j})$ and
$q_j{=}\pi_\phi(\cdot\mid x,f,y_{<j})$.  The loss is the generalized JSD:
\begin{equation}
\mathcal{L}_{\mathrm{distill},j} =
 (1{-}\beta)\,\KL(p_j \| M_j) + \beta\,\KL(q_j \| M_j),
\label{eq:distill-loss}
\end{equation}
where $M_j{=}(1{-}\beta)p_j{+}\beta q_j$ and $\beta{=}0.5$
(symmetric JSD).
Per-token loss is clipped by importance-sampling weight
$\min(\rho,\,c)$ with $c{=}2.0$, where $\rho$ is the token-level
IS ratio between the updated policy and the rollout policy,
correcting for within-mini-batch distribution shift.

\subsection{Collapse Detection and Early Termination}
\label{sec:collapse-detection}

For performance reporting, \emph{collapse} is the first evaluated step
with test mean@16 $<0.01$, using the evaluation frequency above.
Persistence and recovery are separate observations.  Length and entropy
are diagnostics; \emph{reference collapse} describes failure of the teacher
reference, not an accuracy threshold crossing.

The short-horizon sweep has only a final test evaluation.  $M{=}1$
is terminated at step~16 without a test score; $M{=}2$ finishes at
$0.225$.  The step~10 and step~87 entries in Table~\ref{tab:main}
are reported reference-collapse markers, not measured accuracy
threshold crossings.  Figure~\ref{fig:length-tail} retains the step~87
marker for $M{=}2$: its mean response length rises from approximately
2,555 tokens at step~87 to 6,690 at step~88 and remains near the
8,192-token cap thereafter.  Earlier accuracy crossings are undetermined.
Long-horizon $M{=}50$ first crosses the accuracy threshold at step~189
(test $\to0.000$).

%%%%%%%%%%%%%%%%%%%%%%%%%%%%%%%%%%%%%%%%%%%%%%%%%%%%%%%%%%%
\section{Detailed Schedule-Sweep Results}
\label{sec:appendix-details}
%%%%%%%%%%%%%%%%%%%%%%%%%%%%%%%%%%%%%%%%%%%%%%%%%%%%%%%%%%%

\subsection{Additional Metrics}

Table~\ref{tab:full} provides additional metrics for the schedule sweep.

\begin{table*}[h]
\caption{Additional schedule-sweep metrics (100 steps).
Succ = success\_group\_fraction mean;
Ent.\,fin = last-step entropy.
$\dagger$M1 terminated step~16; ent.\ mean over steps 1--16 only.}
\centering
\small
\label{tab:full}
\setlength{\tabcolsep}{4pt}
\begin{tabular}{lcccccccc}
\toprule
\textbf{Setting} &
  \textbf{T\,m@16} &
  \textbf{T\,B@4} &
  \textbf{Ent.\,mn} &
  \textbf{Ent.\,sd} &
  \textbf{Ent.\,fin} &
  \textbf{Rew.\,mn} &
  \textbf{KL\,mn} &
  \textbf{Succ} \\
\midrule
M1$^\dagger$   & N/A    & N/A    & $\sim$1.05 & ---   & 0.492 & ---   & ---   & --- \\
M2             & 0.2250 & 0.2760 & 0.900 & 1.063 & 5.226 & 0.504 & 0.031 & 0.746 \\
M3             & 0.6729 & 0.7396 & 1.715 & 1.167 & 3.663 & 0.607 & 0.085 & 0.773 \\
M4             & 0.6562 & 0.7175 & 1.536 & 1.131 & 4.013 & 0.595 & 0.053 & 0.773 \\
M5             & 0.6533 & 0.7228 & 2.001 & 1.091 & 3.548 & 0.629 & 0.064 & 0.793 \\
M8             & 0.5964 & 0.6291 & 1.413 & 0.713 & 2.641 & 0.597 & 0.004 & 0.681 \\
M10            & 0.6693 & 0.7373 & 0.545 & 0.278 & 0.839 & 0.614 & 0.025 & 0.770 \\
M20            & 0.6286 & 0.6894 & 0.595 & 0.119 & 0.697 & 0.598 & 0.004 & 0.706 \\
M50            & 0.7548 & 0.8273 & 0.499 & 0.048 & 0.416 & 0.629 & 0.005 & 0.816 \\
Fixed          & 0.6967 & 0.7508 & 0.437 & 0.097 & 0.165 & 0.637 & 0.008 & 0.760 \\
EMA $\alpha$=0.1  & 0.6449 & 0.6890 & 1.417 & 0.694 & 2.248 & 0.613 & 0.005 & 0.684 \\
EMA $\alpha$=0.04 & 0.7128 & 0.7815 & 0.727 & 0.142 & 1.864 & 0.665 & 0.022 & 0.792 \\
\bottomrule
\end{tabular}
\end{table*}

%%%%%%%%%%%%%%%%%%%%%%%%%%%%%%%%%%%%%%%%%%%%%%%%%%%%%%%%%%%
\section{CGTR Hyperparameter Values and Selection Rationale}
\label{sec:appendix-cgtr-hparams}
%%%%%%%%%%%%%%%%%%%%%%%%%%%%%%%%%%%%%%%%%%%%%%%%%%%%%%%%%%%

Table~\ref{tab:cgtr-hparams} lists all CGTR gate parameters and
their values used in every experiment reported in this paper.
\emph{No per-dataset tuning was performed}: the same values are
used for Chemistry (long-horizon comparison), Biology, Physics, and ToolUse.

\paragraph{Threshold selection protocol.}
All gate thresholds were determined from the \emph{100-step schedule
sweep} (Section~\ref{sec:results-collapse}--\ref{sec:analysis-fragility}),
which involves only the short-horizon diagnostics on Chemistry.
The long-horizon Chemistry run (300 steps) and all cross-task runs
(Biology, Physics, ToolUse) were conducted \emph{after} the parameters
were fixed, with no further adjustment.
This ordering ensures that the cross-task results constitute a proper
held-out evaluation: the gate thresholds had zero exposure to the
long-horizon or cross-task data at the time of selection.

\begin{table}[t]
\caption{CGTR gate parameters used in all experiments.
Rationale for each value is discussed in the text below.}
\centering
\small
\renewcommand{\arraystretch}{1.12}
\setlength{\tabcolsep}{7pt}
\label{tab:cgtr-hparams}
\begin{tabular}{@{}lcl@{}}
\toprule
\textbf{Parameter} & \textbf{Value} & \textbf{Role} \\
\midrule
$M_{\min}$          & 30   & Minimum isolation period (steps) \\
$w$                 & 5    & Rolling window size (steps) \\
Check interval      & 5    & Gate evaluation frequency (steps) \\
$\delta_r$          & 0.05 & Reward improvement threshold \\
$L_{\max}$          & 3.2  & Length-tail refresh-guard threshold ($\max/\mu$) \\
\bottomrule
\end{tabular}
\end{table}

\paragraph{Selection of $M_{\min}=30$.}
The minimum isolation period was set to be strictly less than the best
fixed-schedule baseline ($M{=}50$), so that CGTR retains the freedom
to refresh more frequently when conditions warrant, while still
exceeding the $M{\le}5$ regime where isolation is too brief to preserve temporal independence
(Section~\ref{sec:results-collapse}).
The value 30 was read directly from the schedule-sweep failure
boundary---a structural property of the model and task class, not
a tuned quantity---and fixed before any long-horizon run.

\paragraph{Selection of $\delta_r=0.05$.}
The reward signal $r_t = \mathbf{1}[\text{correct}]$ has a binary
scale $[0,1]$ with per-window noise $\approx\!\pm0.02$ (estimated
from the rolling variance over the schedule-sweep runs at step~100).
A threshold of $0.05$ places the bar at roughly $2.5\times$ the
noise level, requiring a gain that is unlikely to be a sampling
fluctuation.
This choice is grounded in the noise floor of the reward estimator,
as estimated from the binary-reward Chemistry sweep with the fixed
rollout batch configuration ($n{=}4$).

\begin{figure}[tbp]
\centering
\includegraphics[width=\linewidth]{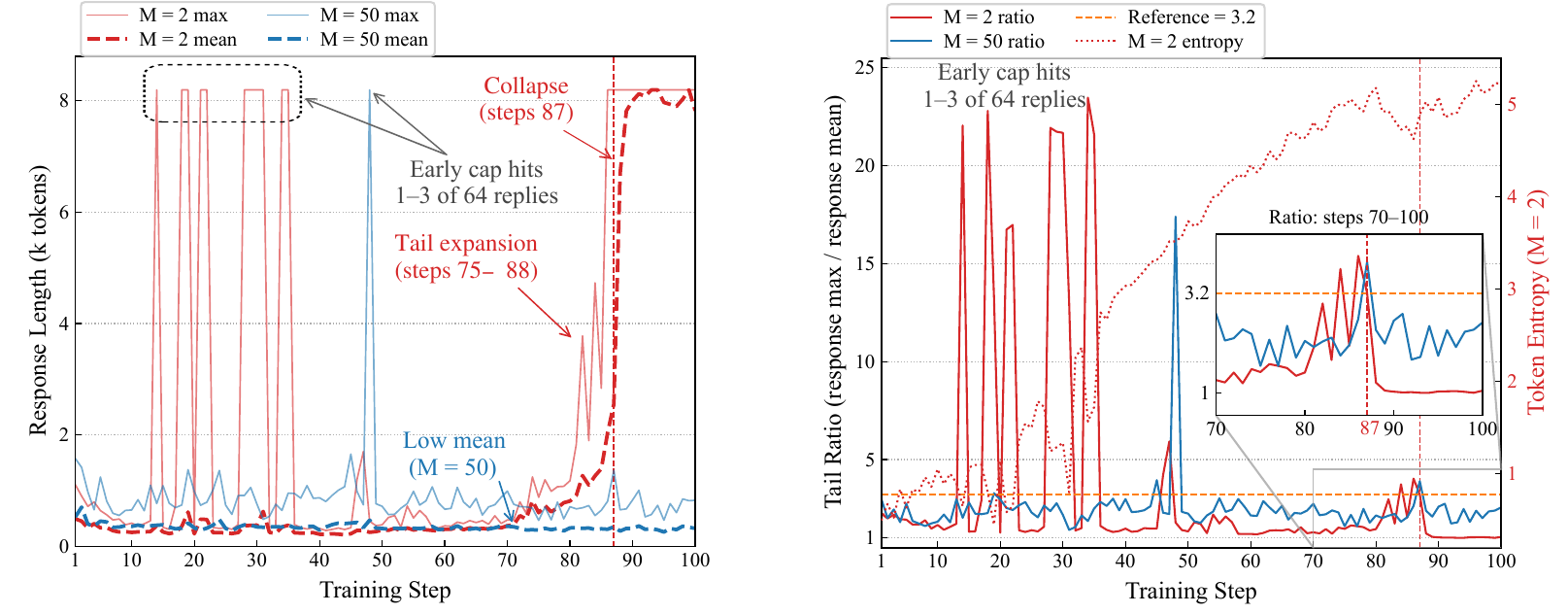}
\caption{%
Response-length dynamics on Chemistry over 100 training steps.
\textbf{(Left)}~Per-step maximum (solid) and mean (dashed) response
lengths for $M{=}2$ and $M{=}50$. Early cap hits involve only $1$--$3$ of $64$ replies, while $M{=}2$
develops sustained late mean-length growth toward the 8,192-token cap.
\textbf{(Right)}~Corresponding maximum-to-mean ratios and $M{=}2$
token entropy (dotted, right axis); the inset magnifies steps~70--100.
Vertical dashed lines mark the reported $M{=}2$ reference-collapse
step~87, and the orange line denotes $L_{\max}{=}3.2$.
}
\label{fig:length-tail}
\end{figure}

\paragraph{Selection of $L_{\max}=3.2$.}
We use $L_{\max}=3.2$ as an empirical refresh-guard setting,
fixed with the other thresholds after the Chemistry diagnostic sweep
and before the long-horizon and cross-task experiments.
The scale-normalized ratio is also sensitive to isolated maxima:
Figure~\ref{fig:length-tail} shows early threshold exceedances when only
$1$--$3$ of $64$ responses reach the token cap while the mean remains
low.  We therefore do not treat $3.2$ as a collapse classifier or claim
that all stable steps lie below it.
The setting instead reflects the scale of the late $M{=}2$ excursion:
over steps~75--88, ratio growth is accompanied by sustained growth in
mean response length, with the ratio exceeding~$4.0$ immediately before
the reported step-87 marker.  Setting $L_{\max}=3.2$ makes such unusually
dispersed tails sufficient to veto a candidate refresh.  Because all
three gates must pass, an isolated exceedance only delays a refresh; it
does not label the run as unstable.
The Biology sensitivity analysis
(Appendix~\ref{sec:appendix-sensitivity}) tests
$L_{\max}\in\{2.8,3.2,4.0\}$, with final scores
$0.480$, $0.485$, and $0.469$, respectively, and no observed collapse.
These results support the combined rule within the tested range,
rather than a standalone stability boundary at $3.2$.

%%%%%%%%%%%%%%%%%%%%%%%%%%%%%%%%%%%%%%%%%%%%%%%%%%%%%%%%%%%
\section{Hyperparameter Sensitivity Analysis}
\label{sec:appendix-sensitivity}
%%%%%%%%%%%%%%%%%%%%%%%%%%%%%%%%%%%%%%%%%%%%%%%%%%%%%%%%%%%

To provide the strongest possible evidence that the gate thresholds
generalize across tasks, we conduct the sensitivity sweep on
\textbf{Biology}---a task whose parameters were never involved in
any threshold selection.
We vary each gate threshold independently (300 steps) while holding
the other two at their default values.
Table~\ref{tab:sensitivity} reports test\,fin and \#Ref for each setting.

\begin{table}[h]
\caption{%
Hyperparameter sensitivity on Biology (300 steps).
Bolded rows are the default (selected from Chemistry schedule sweep only).
Test\,fin = final mean@16; \#Ref = refreshes; Coll.\ follows
Appendix~\ref{sec:collapse-detection} (--- = none observed).
}
\centering
\small
\setlength{\tabcolsep}{4pt}
\label{tab:sensitivity}
\begin{tabular}{lrccc}
\toprule
\textbf{Param.} & \textbf{Value} & \textbf{Test\,fin} & \textbf{\#Ref} & \textbf{Coll.} \\
\midrule
\multirow{3}{*}{$M_{\min}$}
  & 20            & 0.471 & 3 & --- \\
  & \textbf{30}   & \textbf{0.485} & \textbf{2} & \textbf{---} \\
  & 40            & 0.479 & 2 & --- \\
\midrule
\multirow{3}{*}{$\delta_r$}
  & 0.02          & 0.463 & 4 & --- \\
  & \textbf{0.05} & \textbf{0.485} & \textbf{2} & \textbf{---} \\
  & 0.10          & 0.474 & 1 & --- \\
\midrule
\multirow{3}{*}{$L_{\max}$}
  & 2.8           & 0.480 & 2 & --- \\
  & \textbf{3.2}  & \textbf{0.485} & \textbf{2} & \textbf{---} \\
  & 4.0           & 0.469 & 3 & --- \\
\bottomrule
\end{tabular}
\end{table}

Across all nine variants test\,fin ranges from $0.463$ to $0.485$,
a spread of $0.022$.
Seven of nine variants match or exceed the EMA-04 baseline
($0.471$); the two exceptions ($\delta_r{=}0.02$: $0.463$;
$L_{\max}{=}4.0$: $0.469$) fall within $0.008$ of EMA and correspond
to configurations that are deliberately suboptimal by design
(over-frequent refreshes and a loose length guard, respectively).
Crucially, \textbf{no variant collapses}, whereas $M{=}50$ collapses
at step~130 on Biology with the same setup---the safety guarantee
is robust across the entire parameter range tested.
These results confirm that the default thresholds, derived solely from
Chemistry diagnostics, transfer to a held-out task without requiring
retuning, and that the performance plateau is broad rather than sharp.

%%%%%%%%%%%%%%%%%%%%%%%%%%%%%%%%%%%%%%%%%%%%%%%%%%%%%%%%%%%
\section{Lag Alignment Derivation}
\label{sec:appendix-lag}
%%%%%%%%%%%%%%%%%%%%%%%%%%%%%%%%%%%%%%%%%%%%%%%%%%%%%%%%%%%

For a hard refresh with interval $M$, the teacher parameters are
copied from the student at every $M$-th step and held frozen in
between.  At any step within a refresh cycle, the teacher parameters
were set $k$ steps ago, where $k$ is uniformly distributed over
$\{1, 2, \ldots, M\}$.  The average age of the teacher parameters
is therefore $(M+1)/2$ steps, and the mean lag relative to the
\emph{current} student is $(M-1)/2$ steps.

For EMA with rate $\alpha$, the teacher at step $t$ is a weighted
average of all past student snapshots with exponentially decaying
weights:
\begin{equation}
\phi_t = \sum_{k=0}^{\infty} \alpha(1-\alpha)^k \theta_{t-k}.
\end{equation}
The mean age of this exponential weighting is $(1-\alpha)/\alpha$
steps.  Setting the two mean ages equal gives the \emph{lag alignment
condition}:
\begin{equation}
\frac{1-\alpha}{\alpha} = \frac{M-1}{2}
\quad\Longrightarrow\quad
\alpha \approx \frac{2}{M+1}.
\label{eq:lag-align-appendix}
\end{equation}
For $M{=}50$, this yields $\alpha \approx 2/51 \approx 0.039$; we use
$\alpha = 0.04$ as the lag-aligned EMA baseline throughout the paper.
If lag alone explained the performance difference between $M{=}50$ and
EMA, then lag-aligned EMA should match sparse hard refresh across all
training horizons.  As shown in Section~\ref{sec:results-m50}, it does
not---the advantage of hard refresh is structural (isolation periods)
rather than attributable to teacher age.

%%%%%%%%%%%%%%%%%%%%%%%%%%%%%%%%%%%%%%%%%%%%%%%%%%%%%%%%%%%
\section{Terminology}
\label{sec:appendix-concepts}
%%%%%%%%%%%%%%%%%%%%%%%%%%%%%%%%%%%%%%%%%%%%%%%%%%%%%%%%%%%

We collect here the key terms used throughout the paper.

\paragraph{Temporal independence.}
A teacher is \emph{temporally independent} when its historical state
retains predictive information not already present in the current
student's on-policy distribution.  As $M \to 1$ or $\alpha \to 1$,
temporal independence vanishes and the distillation objective becomes
self-confirming.

\paragraph{Isolation period.}
In periodic hard refresh, the steps between two consecutive refreshes
form an \emph{isolation period} during which the teacher is completely
frozen.  Any student drift that occurs during this period is not absorbed
by the teacher.  There is no isolation period in EMA.
An isolation period prevents immediate tracking of student fluctuations,
thereby helping preserve temporal independence.  Very short isolation,
however, may be insufficient.

\paragraph{Teacher contamination.}
When the teacher tracks the student closely (low $M$ or high $\alpha$),
it absorbs the student's transient noise, errors, and distribution
shifts.  The distillation objective then reinforces rather than corrects
these deviations.  We call this \emph{teacher contamination}.  In EMA,
contamination is bounded per step by $\alpha$ but accumulates over long
horizons; in hard refresh, a single ill-timed update can contaminate
the teacher fully and irreversibly.

\paragraph{Refresh shock.}
At each hard refresh, the teacher parameters jump from $\phi_{t-1}$ to
$\theta_t$.  The magnitude of this jump, $\|\theta_t - \phi_{t-1}\|$,
characterizes whether the refresh is \emph{corrective} (the student has
improved during the isolation period) or \emph{contaminating} (the
student has drifted or collapsed).  A corrective refresh is followed
by a transient KL spike as the student re-adapts; a contaminating
refresh shows no such spike.

\paragraph{Length-tail risk.}
The distribution of generated sequence lengths is heavy-tailed under
reinforcement-style training.  Its tail can expand before abrupt changes
appear in mean entropy or reward.  We track \texttt{seqlen\_max} and
\texttt{seqlen\_mean} as early-warning diagnostics.

%%%%%%%%%%%%%%%%%%%%%%%%%%%%%%%%%%%%%%%%%%%%%%%%%%%%%%%%%%%
\section{Multi-Seed Verification}
\label{sec:appendix-multiseed}
%%%%%%%%%%%%%%%%%%%%%%%%%%%%%%%%%%%%%%%%%%%%%%%%%%%%%%%%%%%

We repeat the 300-step comparison of $M{=}50$, EMA $\alpha{=}0.04$,
and CGTR on Chemistry and Biology over three seeds in total:
the original run (seed~0) and two additional runs (seeds~1--2).
The training setup follows Appendix~\ref{sec:appendix-repro}, with
the CGTR hyperparameters in Appendix~\ref{sec:appendix-cgtr-hparams}
held fixed without retuning.
Evaluation is performed every step on Chemistry and every 10 steps
on Biology.

Table~\ref{tab:multiseed} reports the mean and standard deviation
across seeds for final and peak test mean@16, together with the
number of runs that collapse and their collapse-time statistics.
The original results in Tables~\ref{tab:phase3}
and~\ref{tab:generalization} are single-run measurements;
the values here summarize all three seeds.

\begin{table}[H]
\caption{Multi-seed comparison on Chemistry and Biology (300 steps,
three seeds per task and schedule).
Test\,fin/peak = final / peak test mean@16, reported as mean$\pm$std
across seeds.
Collapsed = runs meeting Appendix~\ref{sec:collapse-detection}'s
accuracy criterion / total runs.
Collapse step = mean$\pm$std of the first detected collapse step
among collapsed runs, with the observed range in parentheses;
--- indicates no observed collapse within 300 steps.}
\centering
\small
\renewcommand{\arraystretch}{1.12}
\setlength{\tabcolsep}{3pt}
\begin{tabular}{llcccc}
\toprule
\textbf{Task} & \textbf{Sched.} &
  \textbf{Test\,fin} & \textbf{Test\,peak} &
  \textbf{Collapsed} & \shortstack{\textbf{Collapse step} (range)} \\
\midrule
\multirow{3}{*}{Chemistry}
  & $M{=}50$ & $0.000\pm0.000$ & $0.752\pm0.009$ & 3/3 &
    \begin{tabular}[c]{@{}c@{}}$208\pm21$ (189--230)\end{tabular} \\
  & EMA-04 & $0.711\pm0.012$ & $0.795\pm0.009$ & 0/3 & --- \\
  & CGTR & $\mathbf{0.805\pm0.007}$ & $\mathbf{0.809\pm0.005}$ & 0/3 & --- \\
\midrule
\multirow{3}{*}{Biology}
  & $M{=}50$ & $0.000\pm0.000$ & $0.554\pm0.006$ & 3/3 &
    \begin{tabular}[c]{@{}c@{}}$140\pm26$ (120--170)\end{tabular} \\
  & EMA-04 & $0.471\pm0.008$ & $0.567\pm0.006$ & 0/3 & --- \\
  & CGTR & $\mathbf{0.482\pm0.003}$ & $\mathbf{0.573\pm0.005}$ & 0/3 & --- \\
\bottomrule
\end{tabular}
\label{tab:multiseed}
\end{table}

$M{=}50$ collapses in all three seeds on both tasks, although the
observed collapse time varies from 189 to 230 on Chemistry and
from 120 to 170 on Biology.
CGTR and EMA have no observed collapse in these runs, with CGTR
achieving the highest mean final score on both tasks.
These results support repeatability across the tested seeds;
zero observed collapses in three runs is not a general stability guarantee.

%%%%%%%%%%%%%%%%%%%%%%%%%%%%%%%%%%%%%%%%%%%%%%%%%%%%%%%%%%%
\section{Frozen-Teacher Controls}
\label{sec:appendix-fixed-controls}
%%%%%%%%%%%%%%%%%%%%%%%%%%%%%%%%%%%%%%%%%%%%%%%%%%%%%%%%%%%

We add two frozen-teacher controls on Chemistry for 300 steps.
\textbf{Fixed (init)} holds the pretrained teacher fixed throughout
training. \textbf{Fixed@100} uses the teacher snapshot active in the
$M{=}50$ run at step~100 and freezes it through step~300.
The latter is a freeze-after-warmup control without CGTR's
state-dependent gating.

\begin{table}[H]
\caption{Frozen-teacher controls on Chemistry (300 steps, single run).
Fixed (init) never refreshes the pretrained teacher; Fixed@100 freezes
the $M{=}50$ teacher snapshot from step~100 through step~300.
Test\,fin/peak = final / peak test mean@16;
Ent\,fin = final token entropy; Coll.\ follows
Appendix~\ref{sec:collapse-detection} (--- = none observed).}
\centering
\small
\setlength{\tabcolsep}{5pt}
\begin{tabular}{lcccc}
\toprule
\textbf{Schedule} & \textbf{Test\,fin} & \textbf{Test\,peak} &
  \textbf{Ent\,fin} & \textbf{Coll.} \\
\midrule
Fixed (init) & 0.725 & 0.738 & 0.158 & --- \\
Fixed@100 & 0.765 & 0.768 & 0.420 & --- \\
CGTR & \textbf{0.806} & \textbf{0.809} & 0.780 & --- \\
\bottomrule
\end{tabular}
\label{tab:fixed-controls}
\end{table}

Both frozen-teacher controls avoid collapse, but their final scores
remain below CGTR: $0.725$ for Fixed (init), $0.765$ for Fixed@100,
and $0.806$ for CGTR.
The $0.041$ difference between Fixed@100 and CGTR shows that the tested
freeze-after-warmup control does not reproduce CGTR's final performance.
Because this is a single-run comparison at one selected freeze time,
it does not establish superiority over every possible frozen snapshot.

\end{document}